\pdfoutput=1
\documentclass[akbc,twoside,11pt,lettersize]{article}
\usepackage{akbc}
\akbcheading
\ShortHeadings{Autoregressive Modeling of TKGs in Curvature-Variable Hyperbolic Spaces}{Sohn, Ma, \& Chen}

\usepackage{latexsym}

\usepackage[T1]{fontenc}
\usepackage[utf8]{inputenc}
\usepackage{microtype}

\usepackage{xspace}
\usepackage{amsmath,amssymb}
\usepackage{graphicx,multirow}
\usepackage{pgfplots}
\pgfplotsset{compat=1.17}
\usepackage{wrapfig}

\newcommand{\model}{\texttt{HyperVC}\xspace}

\newcommand{\poincare}{Poincar\'e\xspace}

\newcommand{\stitle}[1]{\vspace{0.3em}\noindent{\bf #1}}

\finalcopy

\begin{document}

\title{Bending the Future: Autoregressive Modeling of Temporal Knowledge Graphs in Curvature-Variable Hyperbolic Spaces}

\author{\name Jihoon Sohn \email jihoonso@usc.edu \\
       \addr Department of Mathematics\\
       University of Southern California
       \AND
       \name Mingyu Derek Ma \email ma@cs.ucla.edu \\
       \addr Department of Computer Science\\
       University of California, Los Angeles
       \AND
       \name Muhao Chen \email muhaoche@usc.edu \\
       \addr Department of Computer Science\\
       University of Southern California}

\maketitle

\begin{abstract}
Recently there is an increasing scholarly interest in time-varying knowledge graphs, or temporal knowledge graphs (TKG). Previous research suggests diverse approaches to TKG reasoning that uses historical information. However, less attention has been given to the hierarchies within such information at different timestamps. Given that TKG is a sequence of knowledge graphs based on time, the chronology in the sequence derives hierarchies between the graphs. Furthermore, each knowledge graph has its hierarchical level which may differ from one another. To address these hierarchical characteristics in TKG, we propose \model, which utilizes hyperbolic space that better encodes the hierarchies than Euclidean space. The chronological hierarchies between knowledge graphs at different timestamps are represented by embedding the knowledge graphs as vectors in a common hyperbolic space. Additionally, diverse hierarchical levels of knowledge graphs are represented by adjusting the curvatures of hyperbolic embeddings of their entities and relations. Experiments on four benchmark datasets show substantial improvements, especially on the datasets with higher hierarchical levels.
\end{abstract}

\section{Introduction}

Knowledge graphs (KGs) have been the backbone of many knowledge-driven AI applications \cite{zhang2016collaborative,wang2018dkn,liu-etal-2018-entity,huang2019knowledge},
where factual knowledge about the real world are described as graphs of entities (nodes) and relations (edges).
Yet, facts in real-world are varying over time instead of being persistently unchanged.
For example, $\left< \textit{JoeBiden}, \textit{IsPresidentOf}, \textit{U.S.} \right>$ has been true only for a year-long time, since January 20th, 2021. 
We do not know when this fact will turn to false--he may run for another presidential term, or he may not--but we know that this will eventually turn to false at some time.
Hence, it is meaningful to represent such facts that are dynamically changing over time using a temporal knowledge graph (TKG) in the form of $\left< \textit{subject}, \textit{relation}, \textit{object}, \textit{time} \right>$.
TKG representation has numerous downstream applications including event prediction \citep{luo2020dynamic,deng2020dynamic}, transaction recommendation \cite{ren2019repeatnet} and schema induction \citep{zhang2020analogous}.

The main purpose of TKG reasoning is to forecast future events or facts \cite{jin-etal-2020-recurrent,Zhu_Chen_Fan_Cheng_Zhang_2021,trivedi2017know,trivedi2019dyrep}. 
To precisely define the task, consider a TKG where events lie in a temporal interval $[t_0, t_T]$. 
Instead of predicting events at timestamps $t \in [t_0, t_T]$, this task, also known as \textit{extrapolation}, aims to predict new facts at a timestamp $t > t_T$.
To tackle this task, recent studies have proposed several approaches \citep{jin-etal-2020-recurrent, Zhu_Chen_Fan_Cheng_Zhang_2021, he2021hipn}.
\citet{jin-etal-2020-recurrent} proposed an autoregressive approach while \citet{Zhu_Chen_Fan_Cheng_Zhang_2021} utilized the copy-generation mechanism. 
\citet{he2021hipn} focused on both structural and temporal perspectives.
However, there are still a few uncovered challenges.
Given that a TKG is a time series of KGs, it is natural that hierarchies can be chronologically derived from TKGs.
In particular, one event may evolve into several relevant subevents \citep{suris2021hyperfuture}, forming hierarchies that represent different paths of evolution among KGs. 
Little attention has been given to incorporating such hierarchical structures in previous research. 
Moreover, each KG at one timestamp has different numbers of entities and relations, hence different characteristics as a graph.
Note that the characteristics of two graphs at contiguous timestamps are relatively similar compared to those of two graphs at distant timestamps.
Previous studies encoded entities and relations in the Euclidean space, which can not fully address optimizing embedding space for each KG of various graph structures. 
Hyperbolic spaces, often considered as a continuous version of trees, are more advantageous to Euclidean spaces in encoding asymmetric and hierarchical relations \citep{chami2019hyperbolic, liu2019hyperbolic}.
The curvature of hyperbolic space decides the expanding ratio of the space that fits data structures with a certain exponential factor.

To represent hierarchical and chronological properties of events in the TKG, we propose \model (\textbf{Hyper}bolic model with \textbf{V}ariable \textbf{C}urvature), a TKG embedding model in the hyperbolic space instead of a Euclidean spaces. 
\model utilizes hyperbolic spaces in two ways in representations of global information and local information, respectively, of each snapshot of TKG. 
Global representation summarizes global information of a KG at one timestamp. 
To represent hierarchical structures among snapshots, all global representations are embedded in the common hyperbolic space.
On the other hand, local representations focus on local information such as an entity or an entity-relation pair. 
To improve the optimization of the embedding space for each snapshot, \model gives a variety in the curvature of the embedding space of each snapshot to optimize the embedding spaces that represent KGs of various structures.
In other words, distinct structures of KGs are embedded in hyperbolic spaces with different curvatures.
Specifically, it is natural for hyperbolic KG embedding models to efficiently represent KGs where the curvature of embedding space was proportional to how hierarchical the graph is \citep{balazevic2019multi, chami-etal-2020-low}. 
In particular, their analyses with Krackhardt hierarchical scores \citep{krackhardt2014graph} show that the greater the hierarchical score of the data, the better performance at hyperbolic embedding the data showed. 
Hence, by controlling the curvatures of the embedding space, each embedding space of multiple graphs with different hierarchies can be optimized.

\model finds a joint probability distribution of all events in TKG in an autoregressive way, inspired by \citet{jin-etal-2020-recurrent}. Specifically, to learn global and local representations in hyperbolic spaces, \model aggregates the information in the neighborhood using GAT \citep{velickovic2018graph}, encodes facts using hyperbolic RNN \citep{ganea2018hyperbolic}, and decodes as a joint probability distribution of facts. Finally, we infer a curvature of embedding space in future timestamps using a time series model and predict upcoming events using representations of information of graph at hyperbolic space with inferred curvature.

The technical contributions of this work are as follows: 
(1) \model is the first hyperbolic TKG reasoning method that tackles extrapolation task forecasting future events.
(2) Specifically, our method addresses a hierarchy between graphs at different timestamps, which is derived from chronology, and employs it through hyperbolic embedding.
(3) Furthermore, \model applies hyperbolic RNN to deal with representations in hyperbolic spaces and optimizes the curvatures at each timestamp as time series or functions of the hierarchical scores.
(4) \model shows a significant improvement in TKG link prediction task, particularly in data with more hierarchical relations.

\section{Related Works}

\stitle{Temporal KG reasoning}
As discussed by \citet{jin-etal-2020-recurrent}, temporal KG reasoning can be divided into two task settings that aim at predicting facts that are positioned differently on the timeline.
In the \textit{interpolation} setting, the models \citep{jiang2016towards, sadeghian2016temporal, dasgupta2018hyte, garcia-duran-etal-2018-learning, leblay2018deriving, goel2020diachronic, montella-etal-2021-hyperbolic} infer missing facts at the historical timestamps. To do so, \citet{dasgupta2018hyte} projected the entities and relations onto timestamp-specific hyperplanes. \citet{leblay2018deriving} and  \citet{garcia-duran-etal-2018-learning} considered the time as a second relation and integrated times with relations.

On the other hand, in the \textit{extrapolation} setting, the models \citep{jin-etal-2020-recurrent, Zhu_Chen_Fan_Cheng_Zhang_2021, li-etal-2021-search, li2021temporal, he2021hipn, zhou2021sedyt, sun-etal-2021-timetraveler, han-etal-2021-learning-neural} seek to forecast events at unseen (future) timestamp. 
\citet{jin-etal-2020-recurrent} defined a joint probability distribution of all facts in an autoregressive function and \citet{Zhu_Chen_Fan_Cheng_Zhang_2021} developed a time-aware copy-generation mechanism and applied it in TKG embedding. \citet{li-etal-2021-search} and \citet{li2021temporal} utilized graph convolutional networks (GCN) to capture structural dependencies between KGs in adjacent timestamps and \citet{he2021hipn} further considered repetitive perspective of relations. \citet{zhou2021sedyt} proposed a framework that is compatible with most sequence models. \citet{sun-etal-2021-timetraveler} proposed a TKG reasoning model that can handle unseen entities and \citet{han-etal-2021-learning-neural} implemented neural ordinary differential equations to forecast future links on TKGs. 

\stitle{Hyperbolic representation learning}
Data with hierarchical structures can be better represented in the negative-curved hyperbolic space.
Theoretically, this is because the circumference of a hyperbolic space grows exponentially with the radius, which aligns with the size growth of hierarchical data that is also exponential with regard to the level of hierarchies.
Existing works also support this: \citet{nickel2017poincare} proposed a Riemannian optimization method to learn hyperbolic embeddings supervisedly, and \citet{ganea2018hyperbolic} extended neural network operations in the hyperbolic space. Hyperbolic operations in Graph Neural Networks (GNN) using intermediate Euclidean tangent space with differentiable exponential and logarithmic mapping are derived by \citet{liu2019hyperbolic} and \citet{chami2019hyperbolic}. \citet{Dai2021AHG} and \citet{zhang2021lorentzian} further introduced a hyperbolic GCN that less relied on Euclidean tangent space. Hyperbolic representation learning has been applied to tasks such as knowledge graph completion \cite{wang2021mixed,balazevic2019multi}, taxonomy expansion \cite{ma-etal-2021-hyperexpan-taxonomy}, organizational chart induction \cite{chen2019embedding}, event prediction \cite{suris2021hyperfuture}, classification \cite{lopez-strube-2020-fully,chen2020hyperbolic} and knowledge association \cite{sun2020knowledge}. Most related to our work, \citet{han2020dyernie} proposed DyERNIE to use hyperbolic embeddings to capture geometric features of TKGs. \citet{montella-etal-2021-hyperbolic} extended the DyERNIE work and defined the curvature of a Riemannian manifold as the product of both relation and time and shows the helpfulness of the adaptive curvature defined by relations. 

While the two hyperbolic TKG embedding methods tackled the interpolation task, our proposed method tackles the extrapolation task of forecasting future events based on the past. In addition to the differences in the targeted tasks, our method is distinctive from the two earlier methods in terms of model setting and definition. Specifically, in \citet{han2020dyernie}, the entity representations are set to be ``linear'' to the time and curvatures are fixed over time. However, \model optimizes the entity representations through an auto-regressive way and finds the best curvature at each timestamp. Additionally, \citet{montella-etal-2021-hyperbolic} defined the curvature of the Riemannian manifold as a product of two parameters, i.e., the relation-dependent parameter and the time-dependent parameter. While their method may struggle in finding the time-dependent parameter for the future (unseen) timestamp, our method applies to forecasting the future because we train the curvature as a function of times and Krackhardt hierarchical scores.

\section{Hyperbolic Spaces}

\stitle{General property}
A hyperbolic space is a Riemannian space with constant negative curvature, whereas the curvature of a Euclidean space is constantly zero and that of a Spherical space is constantly positive~\cite{iversen1992hyperbolic}. The curvature of a Riemannian space characterizes how the space is locally structured. Particularly, a negative curvature indicates that the volumes grow faster than in the Euclidean space~\cite{cannon1997hyperbolic}.  

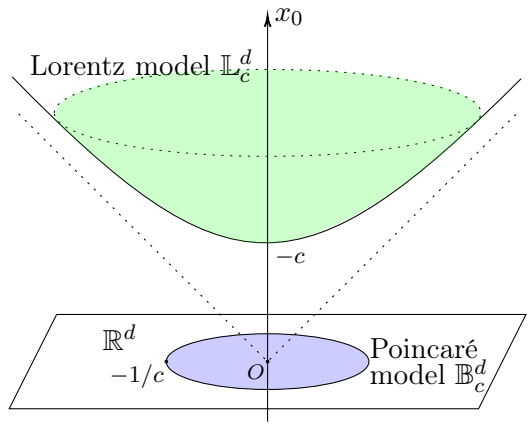
\begin{wrapfigure}{r}{0.5\textwidth}
    \centering
    \vspace{-10pt}
    \begin{tikzpicture}[x=0.35pt,y=0.35pt,yscale=-1,xscale=1]
    \draw  [dash pattern={on 0.84pt off 2.51pt}]  (50,116) .. controls (50,90.04) and (153.2,69) .. (280.5,69) .. controls (407.8,69) and (511,90.04) .. (511,116) .. controls (511,141.96) and (407.8,163) .. (280.5,163) .. controls (153.2,163) and (50,141.96) .. (50,116) -- cycle ;
    \draw    (4.5,77) .. controls (237.5,316) and (317.5,316) .. (554.5,79) ;
    \draw  [dash pattern={on 0.84pt off 2.51pt}]  (11.5,117) -- (280.5,385) -- (550.5,116) ;
    \draw  [fill={rgb, 255:red, 0; green, 0; blue, 255 }  ,fill opacity=0.2 ] (171,385) .. controls (171,368.36) and (220.02,354.88) .. (280.5,354.88) .. controls (340.98,354.88) and (390,368.36) .. (390,385) .. controls (390,401.64) and (340.98,415.13) .. (280.5,415.13) .. controls (220.02,415.13) and (171,401.64) .. (171,385) -- cycle ;
    \draw   (52.5,334) -- (559.75,334) -- (508.5,436) -- (1.25,436) -- cycle ;
    \draw  [draw opacity=0][fill={rgb, 255:red, 0; green, 255; blue, 0 }  ,fill opacity=0.2 ] (280.5,69) .. controls (485.5,71) and (526.5,106) .. (506.5,126) .. controls (496.96,135.54) and (466.28,165.09) .. (426.7,193.72) .. controls (383.28,225.13) and (329.15,255.43) .. (280.5,257) .. controls (187.5,260) and (76.5,147) .. (55.5,128) .. controls (34.5,109) and (75.5,67) .. (280.5,69) -- cycle ;
    \draw    (280.5,450) -- (280.5,14) ;
    \draw [shift={(280.5,12)}, rotate = 450][line width=0.75]    (10.93,-3.29) .. controls (6.95,-1.4) and (3.31,-0.3) .. (0,0) .. controls (3.31,0.3) and (6.95,1.4) .. (10.93,3.29)   ;
    
    \draw[fill=black] (280.5,385) circle[radius=0.5pt] node [anchor=north east] [font=\scriptsize][inner sep=0.9pt]   {$O$};
    \draw (286,260) node [anchor=north west][inner sep=0.75pt] [font=\footnotesize]    {$-c$};
    \draw (171,385) circle[radius=0.5pt] node [anchor=north east][inner sep=0.75pt] [font=\footnotesize]    {$-1/c$};
    \draw (21,45) node [anchor=north west][inner sep=0.75pt]    {Lorentz model $\mathbb{L}_c^d$};
    \draw (380,370) node [anchor=west]   {\poincare}; 
    \draw (380,400) node [anchor=west]    {model $\mathbb{B}_c^d$};
        \draw (285.5,12) node [anchor=west] [inner sep=0.75pt]    {$x_{0}$};
    \draw (150,334) node [anchor=north east] {$\mathbb{R}^d$};
    \end{tikzpicture}
    \caption{$\mathbb{B}_c^d$ is a \poincare ball model (inside of purple ball) that are embedded in $\mathbb{R}^{d+1}$ with $x_0 = 0$ and $\mathbb{L}_c^d$ is a Lorentz hyperboloid model (green hyperboloid) of hyperbolic space.}
    \label{fig:hyperbolic}
    \vspace{-1cm}
\end{wrapfigure}

\stitle{Two models of hyperbolic spaces}
Several models describe hyperbolic spaces~\cite{beltrami1868teoria,cannon1997hyperbolic} and we introduce two models here: the \poincare ball model and the Lorentz hyperboloid model. The \poincare ball model, or simply \poincare model, with curvature $c<0$, is a $d$-dimensional ball $\mathbb{B}_c^d=\{ \mathbf{x} \in \mathbb{R}^d \: | \: \| \mathbf{x} \|^2 < -1/c \}$. 
The Lorentz hyperboloid model (simply Lorentz model) is another $d$-dimensional hyperbolic space defined as $\mathbb{L}_c^d = \{ \mathbf{x}=(x_0, \cdots, x_d) \in \mathbb{R}^{d+1} | \left< \mathbf{x}, \mathbf{x} \right>_\mathbb{L} = c, x_0 >0 \}$, where $c$ is the curvature. $\mathbb{H}_c^d$ denotes a $d$-dimensional hyperbolic space of curvature $c$ regardless of models.
See Figure \ref{fig:hyperbolic}.

\stitle{Basic operations: addition and multiplication}
We introduce addition $\oplus_c$ and multiplication $\otimes_c$ commonly used in neural networks on hyperbolic spaces. In the \poincare model, we use M\"obius addition and M\"obius matrix-vector multiplication. In the Lorentz model, addition and multiplication are performed via the tangent space. For the details, see appendix \ref{appen:hyperoperations}.

\stitle{Hyperbolic RNN}
\label{sec:hyperbolic-rnn}
With the basic hyperbolic operations defined above, we introduce how to generalize a Euclidean RNN to the hyperbolic space~\cite{Ganea2018HyperbolicNN}. Traditional RNN is defined as $h_{t+1}=\varphi\left(W h_{t}+U x_{t}+b\right)$ where $\varphi$ is a pointwise non-linearity, $h_{t}$ is the hidden state of previous unit, $x_{t}$ is the input, and $W$, $U$ and $b$ are model parameters on Euclidean space $\mathbb{E}$. Given $\mathbb{H}_{c}$ as the hyperbolic space modeled by a hyperbolic model (such as \poincare or Lorentz model) with curvature $c$, $\mathcal{M}$ as a manifold with a flat surface of a certain point locally approximated by the Euclidean space $\mathbb{E}$, we generalize it as follows:
\begin{equation*}
h_{t+1}=\varphi^{\otimes_{c}}\left(W \otimes_{c} h_{t} \oplus_{c} U \otimes_{c} x_{t} \oplus_{c} b\right)
\end{equation*}
where $h_{t}, x_{t}, b\in \mathbb{H}_{c}$ and $W, U \in \mathcal{M}$. Note that the input embeddings are in the hyperbolic space.

\section{Method}

In this section, we introduce our method \model\footnote{Codes are available at \url{https://github.com/jhsohn11/HyperVC}.}.

\stitle{Notations and problem definitions}
A TKG contains a set of quadraplets $\left( s, r, o, t\right) \in \mathcal{F} \subset \mathcal{E} \times \mathcal{R} \times \mathcal{E} \times \mathcal{T}$, where $\mathcal{F}$, $\mathcal{E}$, $\mathcal{R}$, $\mathcal{T}$ are the set of valid facts, entities, relations, and timestamps, respectively. $G_t$ denotes the set of facts at timestamp $t$ and thus a TKG can be written as $\{ G_t \}_{t \in \mathcal{T}}$. Our goal is to predict a probability $p(s, r, o, t)$ for each triplet $(s, r_t, o)$ of being contained in $G_t$. Note that among $s$, $r$, and $o$, only the relation $r$ is a time-sensitive variable. To do so, we first assume that $G_t$ depends on previous $m$ snapshot graphs $G_{[t-m, t-1]}$. Inspired by \citet{jin-etal-2020-recurrent}, we decompose the probability $p\left( s, r_t, o | G_{[t-m, t-1]} \right)$ into:
\begin{equation}\label{prob_sro}
p\left( s, r_t, o | G_{[t-m, t-1]} \right) = p\left( s | G_{[t-m, t-1]} \right) \cdot p\left( r_t | s, G_{[t-m, t-1]} \right) \cdot p\left( o | s, r_t, G_{[t-m, t-1]} \right).
\end{equation}
Namely, when we compute the probability of a triplet $(s, r_t, o)$, we first sample a subject entity $s$ using $p\left( s | G_{[t-m, t-1]} \right)$. Next we calculate the probability $p\left( r_t | s, G_{[t-m, t-1]} \right)$ of $r_t$ given $s$ and the previous $m$ timestamps $G_{[t-m, t-1]}$. Finally we compute the probability $p\left( o | s, r_t, G_{[t-m, t-1]} \right)$ of $o$ given $s$, $r_t$, and the previous events $G_{[t-m, t-1]}$.

In the prediction of a missing (future) temporal fact, we infer the missing object entity given $(s, r, ?, t)$ or the missing subject entity given $(?, r, o, t)$. For the former case, the prediction is based on the computation in Equation~\ref{prob_sro} of probability, and for the latter, we compute the probability similarly as Equation~\ref{prob_sro} but sampling the object entity $o$ first instead of $s$. 

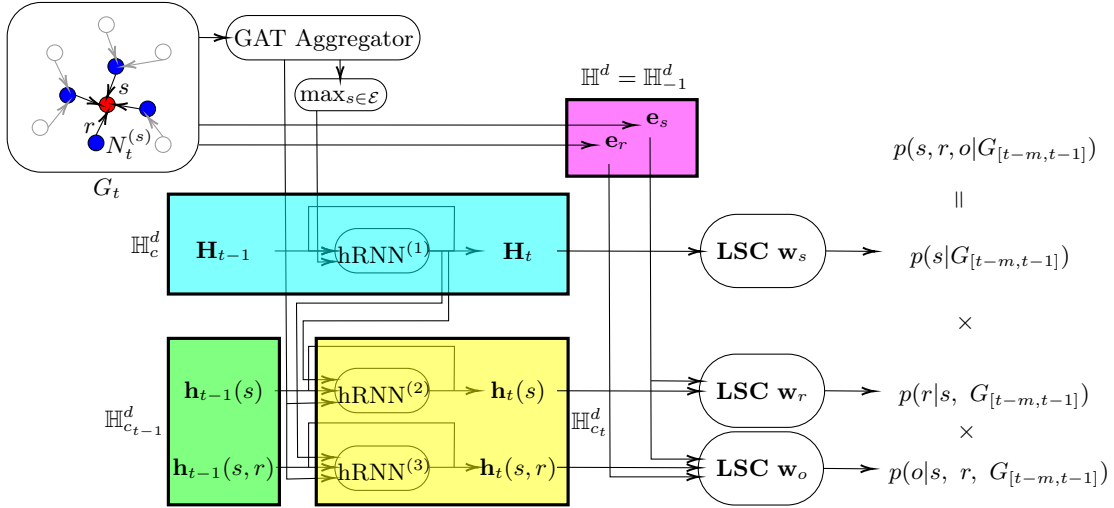
\begin{figure}[t]
    \centering
    \footnotesize
    \begin{tikzpicture}[x=0.42pt,y=0.42pt,yscale=-1,xscale=1]
    \draw   (300,226) .. controls (300,214.95) and (308.95,206) .. (320,206) -- (365,206) .. controls (376.05,206) and (385,214.95) .. (385,226) -- (385,226) .. controls (385,237.05) and (376.05,246) .. (365,246) -- (320,246) .. controls (308.95,246) and (300,237.05) .. (300,226) -- cycle ;
    \draw   (300,422) .. controls (300,410.95) and (308.95,402) .. (320,402) -- (365,402) .. controls (376.05,402) and (385,410.95) .. (385,422) -- (385,422) .. controls (385,433.05) and (376.05,442) .. (365,442) -- (320,442) .. controls (308.95,442) and (300,433.05) .. (300,422) -- cycle ;
    \draw   (300,352) .. controls (300,340.95) and (308.95,332) .. (320,332) -- (365,332) .. controls (376.05,332) and (385,340.95) .. (385,352) -- (385,352) .. controls (385,363.05) and (376.05,372) .. (365,372) -- (320,372) .. controls (308.95,372) and (300,363.05) .. (300,352) -- cycle ;
    \draw    (246.8,421.2) -- (296.8,421.2) ;
    \draw [shift={(298.8,421.2)}, rotate = 180] [color={rgb, 255:red, 0; green, 0; blue, 0 }  ][line width=0.75]    (10.93,-3.29) .. controls (6.95,-1.4) and (3.31,-0.3) .. (0,0) .. controls (3.31,0.3) and (6.95,1.4) .. (10.93,3.29)   ;
    \draw    (245.8,352.2) -- (296.8,352.2) ;
    \draw [shift={(298.8,352.2)}, rotate = 180] [color={rgb, 255:red, 0; green, 0; blue, 0 }  ][line width=0.75]    (10.93,-3.29) .. controls (6.95,-1.4) and (3.31,-0.3) .. (0,0) .. controls (3.31,0.3) and (6.95,1.4) .. (10.93,3.29)   ;
    \draw    (383.8,421.2) -- (426.8,421.2) ;
    \draw [shift={(428.8,421.2)}, rotate = 180] [color={rgb, 255:red, 0; green, 0; blue, 0 }  ][line width=0.75]    (10.93,-3.29) .. controls (6.95,-1.4) and (3.31,-0.3) .. (0,0) .. controls (3.31,0.3) and (6.95,1.4) .. (10.93,3.29)   ;
    \draw    (383.8,226.2) -- (426.8,226.2) ;
    \draw [shift={(428.8,226.2)}, rotate = 180] [color={rgb, 255:red, 0; green, 0; blue, 0 }  ][line width=0.75]    (10.93,-3.29) .. controls (6.95,-1.4) and (3.31,-0.3) .. (0,0) .. controls (3.31,0.3) and (6.95,1.4) .. (10.93,3.29)   ;
    \draw    (383.8,352.2) -- (426.8,352.2) ;
    \draw [shift={(428.8,352.2)}, rotate = 180] [color={rgb, 255:red, 0; green, 0; blue, 0 }  ][line width=0.75]    (10.93,-3.29) .. controls (6.95,-1.4) and (3.31,-0.3) .. (0,0) .. controls (3.31,0.3) and (6.95,1.4) .. (10.93,3.29)   ;
    \draw    (407.34,352.17) -- (407.65,312.2) -- (276.3,312.2) -- (276.3,352.2) ;
    \draw    (245.8,226.2) -- (296.8,226.2) ;
    \draw [shift={(298.8,226.2)}, rotate = 180] [color={rgb, 255:red, 0; green, 0; blue, 0 }  ][line width=0.75]    (10.93,-3.29) .. controls (6.95,-1.4) and (3.31,-0.3) .. (0,0) .. controls (3.31,0.3) and (6.95,1.4) .. (10.93,3.29)   ;
    \draw    (383.8,226.2) -- (407.34,226.17) -- (407.65,186.2) -- (276.3,186.2) -- (276.3,226.2) -- (296.8,226.2) ;
    \draw [shift={(298.8,226.2)}, rotate = 180] [color={rgb, 255:red, 0; green, 0; blue, 0 }  ][line width=0.75]    (10.93,-3.29) .. controls (6.95,-1.4) and (3.31,-0.3) .. (0,0) .. controls (3.31,0.3) and (6.95,1.4) .. (10.93,3.29)   ;
    \draw    (407.34,421.17) -- (407.65,381.2) -- (276.3,381.2) -- (276.3,421.2) ;
    \draw    (402.34,226.17) -- (402.4,289.13) -- (271.4,289.2) -- (271.4,342.2) -- (296.8,342.2) ;
    \draw [shift={(298.8,342.2)}, rotate = 180] [color={rgb, 255:red, 0; green, 0; blue, 0 }  ][line width=0.75]    (10.93,-3.29) .. controls (6.95,-1.4) and (3.31,-0.3) .. (0,0) .. controls (3.31,0.3) and (6.95,1.4) .. (10.93,3.29)   ;
    \draw    (396.3,226.2) -- (396.4,273.2) -- (265.4,273.2) -- (266.4,412.2) -- (296.8,412.2) ;
    \draw [shift={(298.8,412.2)}, rotate = 180] [color={rgb, 255:red, 0; green, 0; blue, 0 }  ][line width=0.75]    (10.93,-3.29) .. controls (6.95,-1.4) and (3.31,-0.3) .. (0,0) .. controls (3.31,0.3) and (6.95,1.4) .. (10.93,3.29)   ;
    \draw  [color={rgb, 255:red, 155; green, 155; blue, 155 }  ,draw opacity=1 ] (27.4,115.01) .. controls (27.4,111.12) and (30.59,107.96) .. (34.53,107.96) .. controls (38.46,107.96) and (41.65,111.12) .. (41.65,115.01) .. controls (41.65,118.91) and (38.46,122.07) .. (34.53,122.07) .. controls (30.59,122.07) and (27.4,118.91) .. (27.4,115.01) -- cycle ;
    \draw  [fill={rgb, 255:red, 0; green, 0; blue, 255 }  ,fill opacity=1 ] (77.74,128.81) .. controls (77.74,124.92) and (80.93,121.76) .. (84.87,121.76) .. controls (88.8,121.76) and (91.99,124.92) .. (91.99,128.81) .. controls (91.99,132.71) and (88.8,135.87) .. (84.87,135.87) .. controls (80.93,135.87) and (77.74,132.71) .. (77.74,128.81) -- cycle ;
    \draw  [color={rgb, 255:red, 155; green, 155; blue, 155 }  ,draw opacity=1 ] (40.57,47.55) .. controls (40.57,43.66) and (43.76,40.5) .. (47.69,40.5) .. controls (51.63,40.5) and (54.82,43.66) .. (54.82,47.55) .. controls (54.82,51.45) and (51.63,54.6) .. (47.69,54.6) .. controls (43.76,54.6) and (40.57,51.45) .. (40.57,47.55) -- cycle ;
    \draw  [fill={rgb, 255:red, 0; green, 0; blue, 255 }  ,fill opacity=1 ] (95.55,59.82) .. controls (95.55,55.92) and (98.74,52.76) .. (102.68,52.76) .. controls (106.61,52.76) and (109.8,55.92) .. (109.8,59.82) .. controls (109.8,63.71) and (106.61,66.87) .. (102.68,66.87) .. controls (98.74,66.87) and (95.55,63.71) .. (95.55,59.82) -- cycle ;
    \draw  [fill={rgb, 255:red, 0; green, 0; blue, 255 }  ,fill opacity=1 ] (124.21,98.15) .. controls (124.21,94.25) and (127.4,91.1) .. (131.33,91.1) .. controls (135.27,91.1) and (138.46,94.25) .. (138.46,98.15) .. controls (138.46,102.04) and (135.27,105.2) .. (131.33,105.2) .. controls (127.4,105.2) and (124.21,102.04) .. (124.21,98.15) -- cycle ;
    \draw  [fill={rgb, 255:red, 0; green, 0; blue, 255 }  ,fill opacity=1 ] (52.18,85.88) .. controls (52.18,81.99) and (55.37,78.83) .. (59.31,78.83) .. controls (63.24,78.83) and (66.43,81.99) .. (66.43,85.88) .. controls (66.43,89.78) and (63.24,92.94) .. (59.31,92.94) .. controls (55.37,92.94) and (52.18,89.78) .. (52.18,85.88) -- cycle ;
    \draw  [color={rgb, 255:red, 155; green, 155; blue, 155 }  ,draw opacity=1 ] (87.03,22.25) .. controls (87.03,18.36) and (90.22,15.2) .. (94.16,15.2) .. controls (98.09,15.2) and (101.28,18.36) .. (101.28,22.25) .. controls (101.28,26.15) and (98.09,29.31) .. (94.16,29.31) .. controls (90.22,29.31) and (87.03,26.15) .. (87.03,22.25) -- cycle ;
    \draw  [color={rgb, 255:red, 155; green, 155; blue, 155 }  ,draw opacity=1 ] (138.15,46.78) .. controls (138.15,42.89) and (141.34,39.73) .. (145.27,39.73) .. controls (149.21,39.73) and (152.4,42.89) .. (152.4,46.78) .. controls (152.4,50.68) and (149.21,53.84) .. (145.27,53.84) .. controls (141.34,53.84) and (138.15,50.68) .. (138.15,46.78) -- cycle ;
    \draw  [color={rgb, 255:red, 155; green, 155; blue, 155 }  ,draw opacity=1 ] (138.15,130.35) .. controls (138.15,126.45) and (141.34,123.29) .. (145.27,123.29) .. controls (149.21,123.29) and (152.4,126.45) .. (152.4,130.35) .. controls (152.4,134.24) and (149.21,137.4) .. (145.27,137.4) .. controls (141.34,137.4) and (138.15,134.24) .. (138.15,130.35) -- cycle ;
    \draw  [fill={rgb, 255:red, 255; green, 0; blue, 0 }  ,fill opacity=1 ] (87.81,94.32) .. controls (87.81,90.42) and (91,87.26) .. (94.93,87.26) .. controls (98.87,87.26) and (102.06,90.42) .. (102.06,94.32) .. controls (102.06,98.21) and (98.87,101.37) .. (94.93,101.37) .. controls (91,101.37) and (87.81,98.21) .. (87.81,94.32) -- cycle ;
    \draw    (66.43,85.88) -- (85.95,93.58) ;
    \draw [shift={(87.81,94.32)}, rotate = 201.53] [color={rgb, 255:red, 0; green, 0; blue, 0 }  ][line width=0.75]    (10.93,-3.29) .. controls (6.95,-1.4) and (3.31,-0.3) .. (0,0) .. controls (3.31,0.3) and (6.95,1.4) .. (10.93,3.29)   ;
    \draw    (84.87,121.76) -- (94.05,103.16) ;
    \draw [shift={(94.93,101.37)}, rotate = 116.28] [color={rgb, 255:red, 0; green, 0; blue, 0 }  ][line width=0.75]    (10.93,-3.29) .. controls (6.95,-1.4) and (3.31,-0.3) .. (0,0) .. controls (3.31,0.3) and (6.95,1.4) .. (10.93,3.29)   ;
    \draw    (124.21,98.15) -- (104.03,94.66) ;
    \draw [shift={(102.06,94.32)}, rotate = 9.82] [color={rgb, 255:red, 0; green, 0; blue, 0 }  ][line width=0.75]    (10.93,-3.29) .. controls (6.95,-1.4) and (3.31,-0.3) .. (0,0) .. controls (3.31,0.3) and (6.95,1.4) .. (10.93,3.29)   ;
    \draw [color={rgb, 255:red, 155; green, 155; blue, 155 }  ,draw opacity=1 ]   (145.27,123.29) -- (132.56,106.79) ;
    \draw [shift={(131.33,105.2)}, rotate = 52.39] [color={rgb, 255:red, 155; green, 155; blue, 155 }  ,draw opacity=1 ][line width=0.75]    (10.93,-3.29) .. controls (6.95,-1.4) and (3.31,-0.3) .. (0,0) .. controls (3.31,0.3) and (6.95,1.4) .. (10.93,3.29)   ;
    \draw [color={rgb, 255:red, 155; green, 155; blue, 155 }  ,draw opacity=1 ]   (145.27,53.84) -- (111.78,59.49) ;
    \draw [shift={(109.8,59.82)}, rotate = 350.43] [color={rgb, 255:red, 155; green, 155; blue, 155 }  ,draw opacity=1 ][line width=0.75]    (10.93,-3.29) .. controls (6.95,-1.4) and (3.31,-0.3) .. (0,0) .. controls (3.31,0.3) and (6.95,1.4) .. (10.93,3.29)   ;
    \draw    (102.68,66.87) -- (95.64,85.39) ;
    \draw [shift={(94.93,87.26)}, rotate = 290.8] [color={rgb, 255:red, 0; green, 0; blue, 0 }  ][line width=0.75]    (10.93,-3.29) .. controls (6.95,-1.4) and (3.31,-0.3) .. (0,0) .. controls (3.31,0.3) and (6.95,1.4) .. (10.93,3.29)   ;
    \draw [color={rgb, 255:red, 155; green, 155; blue, 155 }  ,draw opacity=1 ]   (41.65,115.01) -- (58.06,94.5) ;
    \draw [shift={(59.31,92.94)}, rotate = 128.65] [color={rgb, 255:red, 155; green, 155; blue, 155 }  ,draw opacity=1 ][line width=0.75]    (10.93,-3.29) .. controls (6.95,-1.4) and (3.31,-0.3) .. (0,0) .. controls (3.31,0.3) and (6.95,1.4) .. (10.93,3.29)   ;
    \draw [color={rgb, 255:red, 155; green, 155; blue, 155 }  ,draw opacity=1 ]   (47.69,54.6) -- (58.44,77.03) ;
    \draw [shift={(59.31,78.83)}, rotate = 244.38] [color={rgb, 255:red, 155; green, 155; blue, 155 }  ,draw opacity=1 ][line width=0.75]    (10.93,-3.29) .. controls (6.95,-1.4) and (3.31,-0.3) .. (0,0) .. controls (3.31,0.3) and (6.95,1.4) .. (10.93,3.29)   ;
    \draw [color={rgb, 255:red, 155; green, 155; blue, 155 }  ,draw opacity=1 ]   (94.16,29.31) -- (102,50.88) ;
    \draw [shift={(102.68,52.76)}, rotate = 250.04] [color={rgb, 255:red, 155; green, 155; blue, 155 }  ,draw opacity=1 ][line width=0.75]    (10.93,-3.29) .. controls (6.95,-1.4) and (3.31,-0.3) .. (0,0) .. controls (3.31,0.3) and (6.95,1.4) .. (10.93,3.29)   ;
    \draw   (4.84,32.81) .. controls (4.84,15.91) and (18.54,2.21) .. (35.44,2.21) -- (146.8,2.21) .. controls (163.7,2.21) and (177.4,15.91) .. (177.4,32.81) -- (177.4,124.6) .. controls (177.4,141.5) and (163.7,155.2) .. (146.8,155.2) -- (35.44,155.2) .. controls (18.54,155.2) and (4.84,141.5) .. (4.84,124.6) -- cycle ;
    \draw   (202,34) .. controls (202,22.95) and (210.95,14) .. (222,14) -- (356.6,14) .. controls (367.65,14) and (376.6,22.95) .. (376.6,34) -- (376.6,34) .. controls (376.6,45.05) and (367.65,54) .. (356.6,54) -- (222,54) .. controls (210.95,54) and (202,45.05) .. (202,34) -- cycle ;
    \draw   (264,85.4) .. controls (264,77.45) and (270.45,71) .. (278.4,71) -- (332.2,71) .. controls (340.15,71) and (346.6,77.45) .. (346.6,85.4) -- (346.6,85.4) .. controls (346.6,93.35) and (340.15,99.8) .. (332.2,99.8) -- (278.4,99.8) .. controls (270.45,99.8) and (264,93.35) .. (264,85.4) -- cycle ;
    \draw    (283.6,99.8) -- (284.55,236.2) -- (296.6,235.86) ;
    \draw [shift={(298.6,235.8)}, rotate = 178.37] [color={rgb, 255:red, 0; green, 0; blue, 0 }  ][line width=0.75]    (10.93,-3.29) .. controls (6.95,-1.4) and (3.31,-0.3) .. (0,0) .. controls (3.31,0.3) and (6.95,1.4) .. (10.93,3.29)   ;
    \draw    (254.6,53.8) -- (256.6,363.8) -- (296.6,362.85) ;
    \draw [shift={(298.6,362.8)}, rotate = 178.64] [color={rgb, 255:red, 0; green, 0; blue, 0 }  ][line width=0.75]    (10.93,-3.29) .. controls (6.95,-1.4) and (3.31,-0.3) .. (0,0) .. controls (3.31,0.3) and (6.95,1.4) .. (10.93,3.29)   ;
    \draw    (256.6,363.8) -- (256.6,431.8) -- (296.6,430.85) ;
    \draw [shift={(298.6,430.8)}, rotate = 178.64] [color={rgb, 255:red, 0; green, 0; blue, 0 }  ][line width=0.75]    (10.93,-3.29) .. controls (6.95,-1.4) and (3.31,-0.3) .. (0,0) .. controls (3.31,0.3) and (6.95,1.4) .. (10.93,3.29)   ;
    \draw  [line width=1.2][fill={rgb, 255:red, 0; green, 255; blue, 255 }  ,fill opacity=0.5 ] (150,175) -- (510,175) -- (510,265) -- (150,265) -- cycle ;
    \draw  [line width=1.2][fill={rgb, 255:red, 0; green, 255; blue, 0 }  ,fill opacity=0.5 ] (150,305) -- (250,305) -- (250,455) -- (150,455) -- cycle ;
    \draw  [line width=1.2][fill={rgb, 255:red, 255; green, 255; blue, 0 }  ,fill opacity=0.5 ] (283.2,305) -- (510,305) -- (510,455) -- (283.2,455) -- cycle ;
    \draw  [line width=1.2][fill={rgb, 255:red, 255; green, 0; blue, 255 }  ,fill opacity=0.5 ] (508.8,89.8) -- (626.2,89.8) -- (626.2,158.8) -- (508.8,158.8) -- cycle ;
    \draw    (177.6,33.8) -- (200,33.98) ;
    \draw [shift={(202,34)}, rotate = 180.47] [color={rgb, 255:red, 0; green, 0; blue, 0 }  ][line width=0.75]    (10.93,-3.29) .. controls (6.95,-1.4) and (3.31,-0.3) .. (0,0) .. controls (3.31,0.3) and (6.95,1.4) .. (10.93,3.29)   ;
    \draw    (304.6,53.8) -- (304.78,68.8) ;
    \draw [shift={(304.8,70.8)}, rotate = 269.33] [color={rgb, 255:red, 0; green, 0; blue, 0 }  ][line width=0.75]    (10.93,-3.29) .. controls (6.95,-1.4) and (3.31,-0.3) .. (0,0) .. controls (3.31,0.3) and (6.95,1.4) .. (10.93,3.29)   ;
    \draw   (629.2,226.2) .. controls (629.2,207.97) and (643.97,193.2) .. (662.2,193.2) -- (709,193.2) .. controls (727.23,193.2) and (742,207.97) .. (742,226.2) -- (742,226.2) .. controls (742,244.43) and (727.23,259.2) .. (709,259.2) -- (662.2,259.2) .. controls (643.97,259.2) and (629.2,244.43) .. (629.2,226.2) -- cycle ;
    \draw   (628.2,352.2) .. controls (628.2,333.97) and (642.97,319.2) .. (661.2,319.2) -- (708,319.2) .. controls (726.23,319.2) and (741,333.97) .. (741,352.2) -- (741,352.2) .. controls (741,370.43) and (726.23,385.2) .. (708,385.2) -- (661.2,385.2) .. controls (642.97,385.2) and (628.2,370.43) .. (628.2,352.2) -- cycle ;
    \draw   (627.2,422.2) .. controls (627.2,403.97) and (641.97,389.2) .. (660.2,389.2) -- (707,389.2) .. controls (725.23,389.2) and (740,403.97) .. (740,422.2) -- (740,422.2) .. controls (740,440.43) and (725.23,455.2) .. (707,455.2) -- (660.2,455.2) .. controls (641.97,455.2) and (627.2,440.43) .. (627.2,422.2) -- cycle ;
    \draw    (177.4,130.6) -- (530.8,130.8) ;
    \draw [shift={(532.8,130.8)}, rotate = 180.03] [color={rgb, 255:red, 0; green, 0; blue, 0 }  ][line width=0.75]    (10.93,-3.29) .. controls (6.95,-1.4) and (3.31,-0.3) .. (0,0) .. controls (3.31,0.3) and (6.95,1.4) .. (10.93,3.29)   ;
    \draw    (177.4,112.6) -- (567.8,112.8) ;
    \draw [shift={(569.8,112.8)}, rotate = 180.03] [color={rgb, 255:red, 0; green, 0; blue, 0 }  ][line width=0.75]    (10.93,-3.29) .. controls (6.95,-1.4) and (3.31,-0.3) .. (0,0) .. controls (3.31,0.3) and (6.95,1.4) .. (10.93,3.29)   ;
    \draw    (546.8,147.8) -- (547.8,429.8) -- (625.6,429.8) ;
    \draw [shift={(627.6,429.8)}, rotate = 180] [color={rgb, 255:red, 0; green, 0; blue, 0 }  ][line width=0.75]    (10.93,-3.29) .. controls (6.95,-1.4) and (3.31,-0.3) .. (0,0) .. controls (3.31,0.3) and (6.95,1.4) .. (10.93,3.29)   ;
    \draw    (500,421.2) -- (625.7,421.59) ;
    \draw [shift={(627.7,421.6)}, rotate = 180.17] [color={rgb, 255:red, 0; green, 0; blue, 0 }  ][line width=0.75]    (10.93,-3.29) .. controls (6.95,-1.4) and (3.31,-0.3) .. (0,0) .. controls (3.31,0.3) and (6.95,1.4) .. (10.93,3.29)   ;
    \draw    (500,352.2) -- (626.2,352.59) ;
    \draw [shift={(628.2,352.6)}, rotate = 180.17] [color={rgb, 255:red, 0; green, 0; blue, 0 }  ][line width=0.75]    (10.93,-3.29) .. controls (6.95,-1.4) and (3.31,-0.3) .. (0,0) .. controls (3.31,0.3) and (6.95,1.4) .. (10.93,3.29)   ;
    \draw    (500,226.2) -- (627.2,226.59) ;
    \draw [shift={(629.2,226.6)}, rotate = 180.17] [color={rgb, 255:red, 0; green, 0; blue, 0 }  ][line width=0.75]    (10.93,-3.29) .. controls (6.95,-1.4) and (3.31,-0.3) .. (0,0) .. controls (3.31,0.3) and (6.95,1.4) .. (10.93,3.29)   ;
    \draw    (583.8,123.8) -- (584.8,413.8) -- (625.8,413.8) ;
    \draw [shift={(627.8,413.8)}, rotate = 180] [color={rgb, 255:red, 0; green, 0; blue, 0 }  ][line width=0.75]    (10.93,-3.29) .. controls (6.95,-1.4) and (3.31,-0.3) .. (0,0) .. controls (3.31,0.3) and (6.95,1.4) .. (10.93,3.29)   ;
    \draw    (584.8,343.2) -- (627.2,343.58) ;
    \draw [shift={(629.2,343.6)}, rotate = 180.52] [color={rgb, 255:red, 0; green, 0; blue, 0 }  ][line width=0.75]    (10.93,-3.29) .. controls (6.95,-1.4) and (3.31,-0.3) .. (0,0) .. controls (3.31,0.3) and (6.95,1.4) .. (10.93,3.29)   ;
    \draw    (741,352.2) -- (783.4,352.58) ;
    \draw [shift={(785.4,352.6)}, rotate = 180.52] [color={rgb, 255:red, 0; green, 0; blue, 0 }  ][line width=0.75]    (10.93,-3.29) .. controls (6.95,-1.4) and (3.31,-0.3) .. (0,0) .. controls (3.31,0.3) and (6.95,1.4) .. (10.93,3.29)   ;
    \draw    (740,422.2) -- (782.4,422.58) ;
    \draw [shift={(784.4,422.6)}, rotate = 180.52] [color={rgb, 255:red, 0; green, 0; blue, 0 }  ][line width=0.75]    (10.93,-3.29) .. controls (6.95,-1.4) and (3.31,-0.3) .. (0,0) .. controls (3.31,0.3) and (6.95,1.4) .. (10.93,3.29)   ;
    \draw    (742,226.2) -- (784.4,226.58) ;
    \draw [shift={(786.4,226.6)}, rotate = 180.52] [color={rgb, 255:red, 0; green, 0; blue, 0 }  ][line width=0.75]    (10.93,-3.29) .. controls (6.95,-1.4) and (3.31,-0.3) .. (0,0) .. controls (3.31,0.3) and (6.95,1.4) .. (10.93,3.29)   ;
    
    \draw (345,227) node  {$\mathrm{hRNN}^{(1)}$};
    \draw (345,353) node  {$\mathrm{hRNN}^{(2)}$};
    \draw (345,423) node  {$\mathrm{hRNN}^{(3)}$};
    \draw (200,227) node    {$\mathbf{H}_{t-1}$};
    \draw (200,353) node    {$\mathbf{h}_{t-1}(s)$};
    \draw (200,423) node    {$\mathbf{h}_{t-1}(s,r)$};
    \draw (465,227) node    {$\mathbf{H}_{t}$};
    \draw (465,353) node    {$\mathbf{h}_{t}(s)$};
    \draw (465,423) node    {$\mathbf{h}_{t}(s,r)$};
    
    \draw (110,80) node     {$s$};
    \draw (115,130) node     {$N_{t}^{(s)}$};
    \draw (95,170) node     {$G_{t}$};
    \draw (79,115) node    {$r$};
    \draw (290,36) node  {GAT Aggregator};
    \draw (305,88) node  {$\max_{s\in \mathcal{E}}$};    \draw (541,120.4) node [anchor=north west][inner sep=0.75pt]    {$\mathbf{e}_{r}$};
    \draw (578,100.4) node [anchor=north west][inner sep=0.75pt]    {$\mathbf{e}_{s}$};
    
    \draw (130,220) node {$\mathbb{H}_{c}^{d}$};
    \draw (120,380) node {$\mathbb{H}_{c_{t-1}}^{d}$};
    \draw (530,380) node   {$\mathbb{H}_{c_t}^{d}$};
    \draw (570,70) node {$\mathbb{H}^{d}=\mathbb{H}_{-1}^{d}$};
    
    \draw (685,227) node {\textbf{LSC} $\mathbf{w}_s$};
    \draw (685,353) node {\textbf{LSC} $\mathbf{w}_r$};
    \draw (685,423) node {\textbf{LSC} $\mathbf{w}_o$};

    \draw (816,218.4) node [anchor=north west][inner sep=0.75pt]    {$p( s|G_{[ t-m,t-1]})$};
    \draw (806,343.4) node [anchor=north west][inner sep=0.75pt]    {$p( r|s,\ G_{[ t-m,t-1]})$};
    \draw (797,413.4) node [anchor=north west][inner sep=0.75pt]    {$p( o|s,\ r,\ G_{[ t-m,t-1]})$};
    \draw (857,281.4) node [anchor=north west][inner sep=0.75pt]    {$\times $};
    \draw (858,379.4) node [anchor=north west][inner sep=0.75pt]    {$\times $};
    \draw (857.51,191.07) node [anchor=north west][inner sep=0.75pt]  [rotate=-269.47]  {$=$};
    \draw (801,121.4) node [anchor=north west][inner sep=0.75pt]    {$p( s,r,o|G_{[ t-m,t-1]})$};
    \end{tikzpicture}

    \caption{The overall architecture of \model. Colored boxes are the hyperbolic spaces that contain the hyperbolic representations with designated curvatures. \textbf{LSC} $\mathbf{w}$ refers to the linear softmax classifier parametrized by $\mathbf{w}$. Through the cyan box of the global representations, the green and yellow boxes of the local representations, and the pink box of time-consistent hyperbolic embeddings of entities and relations, we calculate the probability of triplet $(s, r, o)$ at the timestamp $t$.}
    \label{fig:model}
    \vspace{-0.8cm}
\end{figure}

\stitle{Model components} 
As shown in Figure \ref{fig:model}, \model implements \textit{global} representations and \textit{local} representations, as in \citet{jin-etal-2020-recurrent}, to find a joint probability distribution of each event. The global representation $\mathbf{H}_t$ captures the global information of the snapshot $G_t$ at timestamp $t$, which describes preferences as a graph including trend or periodicity. On the other hand, the local representation $\mathbf{h}_t$ concentrates on local information such as a vertex, an edge, or its neighborhood, thus representing more entity-specific or relation-specific behaviors. The two representations capture disjoint features of TKGs. Therefore, we utilize both representations to compute the probability distributions of events.

The key idea of \model is that both representations $\mathbf{H}_t$ and $\mathbf{h}_t$ are embedded in the hyperbolic space, either with a \poincare model, or a Lorentz model. We will compare the performance of models embedded in these two hyperbolic spaces in the result part.

\stitle{Global and local representations}
First of all, the global representations $\mathbf{H}_t$, $t \in \mathcal{T}$ are all embedded in one hyperbolic space $\mathbb{H}^d_{c}$ of dimension $d$ with learnable curvature $c$ of hyperbolic space. This is because there is a hierarchy among contiguous snapshots of events derived from chronological order. Specifically, relevant events (for example, events in causal relations) branch into different paths of evolution depending on previously occurred events, and this arises a hierarchy between KGs at neighboring timestamps \cite{suris2021hyperfuture}. Hence, we embed the global representations into one common hyperbolic space with curvature $c$ to represent hierarchical structures between different snapshot graphs.

Moving on to the second representation, let $\mathbf{h}_t(s)$ be the local representation in the hyperbolic space $\mathbb{H}^d_{c_t}$ of dimension $d$ with curvature $c_t$ for a subject entity $s$ and $\mathbf{h}_t(s,r) \in \mathbb{H}^d_{c_t}$ be the one for a pair of a subject entity $s$ and a relation $r$.
Unlike the global representation, $\mathbf{h}_t$ has curvatures $c_t$ that vary over timestamp $t$ because each snapshot $G_t$ has a different hierarchical level. For example, one graph $G_{t_1}$ at the timestamp $t_1$ may have tree-like (hierarchical) structures, which are represented better in hyperbolic space than Euclidean space. For another graph $G_{t_2}$, they may have less hierarchical relations, whose embedding fits better in Euclidean space. Because the hyperbolic space whose curvature is close to zero is similar to Euclidean space, $c_{t_2}$ will be trained to be closer to zero than $c_{t_1}$. In this way, we can afford any graphs with diverse hierarchical levels by training the learnable curvature $c_t$.

These two separate but complementary representations are defined as:
\begin{align*}
\mathbf{H}_{t} &= \mathrm{hRNN}^{(1)} \left( \mathcal{T}^{c} \left( g'(G_t) \right), \mathbf{H}_{t-1} \right),\\
\mathbf{h}_t (s) &= \mathrm{hRNN}^{(2)} \left( \mathcal{T}^{c_t} \left( g(N_t^{(s)}) \right), \mathcal{T}_{c}^{c_t} (\mathbf{H}_{t}), \mathcal{T}_{c_{t-1}}^{c_t}(\mathbf{h}_{t-1} (s)) \right),\\
\mathbf{h}_t (s,r) &= \mathrm{hRNN}^{(3)} \left( \mathcal{T}^{c_t} \left( g(N_t^{(s)}) \right), \mathcal{T}_{c}^{c_t}(\mathbf{H}_{t}), \mathcal{T}_{c_{t-1}}^{c_t}(\mathbf{h}_{t-1} (s,r))\right),
\end{align*}
where $\mathrm{hRNN}$ are the hyperbolic RNNs as described in Section \ref{sec:hyperbolic-rnn} \citep{ganea2018hyperbolic}, $N_t^{(s)}$ is the subgraph of $G_t$ that contains the entity $s$, $g$ is the neighborhood aggregator using graph attention \citep{velickovic2018graph}, and $g'$ is a max-pooling operation defined as $g'(G_t) = \max \left( \{ g(N_t^{(s)}) \}_{s \in G_t} \right)$ among the aggregated neighborhoods of whole entities $s$ in $G_t$. While the representations $\mathbf{H}_{t-1}$ and $\mathbf{H}_t$ in $\mathrm{hRNN}^{(1)}$ have the same curvature $c$, representations in $\mathrm{hRNN}^{(2)}$ and $\mathrm{hRNN}^{(3)}$ have different curvatures. Hence, we adjust the curvatures of $\mathbf{H}_t$ and $\mathbf{h}_{t-1}$ from $c$ and $c_{t-1}$ to $c_t$ through $\mathcal{T}_c^{c_t}$ and $\mathcal{T}_{c_{t-1}}^{c_t}$, respectively. Finally, as $g(N_t^{(s)})$ and $g'(G_t)$ are Euclidean vectors, we need a transition $\mathcal{T}^{c_t}$ of these vectors to the hyperbolic space of curvature $c_t$.

\stitle{Computations of probabilities}
Based on both representations, we compute the probability $p(o | s, r_t, G_{[t-m, t-1]})$ as follows:
\begin{align*}
p\left( o | s, r_t, G_{[t-m, t-1]} \right) = \mathrm{LSC} \left( [ \mathbf{e}_{s} : \mathbf{e}_{r_t} : \mathcal{T}_{c_{t-1}}^{-1} ( \mathbf{h}_{t-1} (s, r_t) ) ]^\top \cdot \mathbf{w}_{o} \right),
\end{align*}
where $\mathbf{e}_{s}$, $\mathbf{e}_{r} \in \mathbb{H}^d$ are learnable hyperbolic representations of the subject entity $s$ and the relation $r$, respectively, embedded in the hyperbolic space with fixed curvature $-1$. The local representation $\mathbf{h}_{t-1} (s, r_t)$ collects the information of previous snapshots.
However, since $\mathbf{h}_{t-1} \in \mathbb{H}^d_{c_{t-1}}$ while $\mathbf{e}_{s}$, $\mathbf{e}_{r_t} \in \mathbb{H}^d_{-1}$, we adjust the curvature of $\mathbf{h}_{t-1}$ from $c_{t-1}$ to $-1$ through $\mathcal{T}_{c_{t-1}}^{-1}$.
\model tracks and updates the semantic of $(s,r)$ up to $t$ by concatenating both static representations $\mathbf{e}_{s}$, $\mathbf{e}_{r}$ and the time-sensitive representation $\mathbf{h}_{t-1}$. Here, we concatenate hyperbolic representations by appending one at the end of another. After the concatenation, \model computes the probability of the object entity $o$ by passing a linear softmax classifier $\mathrm{LSC}$ parametrized by $\mathbf{w}_{o}$.

Through the similar processes, we compute the two other probabilities $p\left( s | G_{[t-m, t-1]} \right)$ and $p\left( r_t | s, G_{[t-m, t-1]}\right)$ as follows:
\begin{align*}
p\left( s | G_{[t-m, t-1]} \right) &= \mathrm{LSC} \left( \mathbf{H}_{t-1}^\top \cdot \mathbf{w}_{s} \right),\\
p\left( r_t | s, G_{[t-m, t-1]} \right) &= \mathrm{LSC} \left( [ \mathbf{e}_{s} : \mathcal{T}_{c_{t-1}}^{-1} ( \mathbf{h}_{t-1} (s, r_t) ) ]^\top \cdot \mathbf{w}_{r_t} \right),
\end{align*}
where the final probabilities are computed by passing a linear softmax classifier parametrized by $\mathbf{w}_{s}$ and $\mathbf{w}_{r_t}$, respectively.

\stitle{Learnable curvature}
Finally, we train the curvature $c_t$ as a function of two variables: times and Krackhardt hierarchical scores. The real-world data such as daily data or weekly data inevitably has a period. As the curvature of the snapshot at each timestamp is affected by how the data look, there exists a seasonal component in the curvature as well. Inspired by \citet{xu2020temporal}, we decompose the curvature as an additive time series 
\begin{equation}\label{timeseries}
c_t = -\sigma \left( \alpha \sin (\omega t) + (\beta t + \gamma) \right),    
\end{equation}
where each term refers to the seasonal component and the trend component. Since the curvature of hyperbolic space is always negative, we take the ``Softplus'' function $\sigma$. 

On the other hand, the Krackhardt hierarchical score (the formula is described in appendix \ref{appen:hierscore}) also affects the curvature of the hyperbolic space where the graph is embedded. Then the curvature is represented as
\begin{equation}\label{hierscore}
c_t = -\sigma ( f( Khs_{G_t})),    
\end{equation}
where $Khs_{G_t}$ is the Krackhardt hierarchical score of $G_t$.
To find a function $f$ that best describes the relation between hierarchical score and curvature, we experimented with polynomials: a linear function or a quadratic function.

Finally, we experimented with the combination of these two separate approaches as
\begin{equation}\label{combi}
    c_t = -\sigma \left( \alpha \sin (\omega t) + (\beta t + \gamma) + f( Khs_{G_t} )\right).
\end{equation}
The results from these three different approaches were compared in the ablation study in the later section.

\stitle{Learning objective}
Given a subject $s$, a relation $r$, and a timestamp $t$, the model predicts the object entity $o$ based on the probability $p(s, r_t, o)$, considering it as a multi-class classification task where each class corresponds to each entity. Then the loss function $\mathcal{L}$ is:

\begin{equation*}
    \mathcal{L} = - \sum_{(s, r, o, t) \in G} \log p(o|s, r_t)    + \lambda \log p(r_t|s)
\end{equation*}
where $G$ is a set of facts, and $\lambda$ is a hyperparameter that controls each loss term. A similar process works for the subject entity prediction with switched subject and object entities in the loss function.

\stitle{Inference}
\model predicts future events based on past events.
Here, we describe the inference for a missing object, which applies to predicting a missing subject WLOG. Given $s$, $r$, $t$, and the past history of snapshot graphs $G_{[1:t]}$, we predict the object $o$ which has the highest conditional probability.
Inspired by \citet{jin-etal-2020-recurrent}, we use multi-step inference over time. \model samples the events at the next timestamp based on the conditional probability to build a sample graph and we use this in the inference of future timestamps. In other words, from computing $p( s_{t+1}, r_{t+1}, o_{t+1} | G_{[1, t]} )$, or $p(G_{t+1} | G_{[1,t]} )$, we get a sample $\widehat{G}_{t+1}$. Then we can further compute $p(G_{t+2} | \widehat{G}_{t+1}, G_{[1,t]} )$. Through the iterative computation of conditional distribution and sampling from it, we get $p(G_{t+\Delta t}| \widehat{G}_{[t+1, t+\Delta t -1]}, G_{[1,t]})$, the estimate of $p(G_{t+\Delta t} | G_{[1,t]})$.

\section{Experiments}

\stitle{Experimental setup}
We use four representative TKGs datasets, GDELT \citep{leetaru2013gdelt}, ICEWS18 \citep{boschee2015icews}, WIKI \citep{leblay2018deriving}, and YAGO \citep{mahdisoltani2014yago3}. More details about the datasets can be found in Table \ref{table:datastat} and appendix \ref{appen:datasets}.

We split each dataset into three subsets by train, validation, test with the proportion of approximately 80\%, 10\%, 10\%, respectively, in chronological order. Note that three subsets contain disjoint timestamps.
We report two evaluation metrics for extrapolated link prediction, including Mean Reciprocal Rank (MRR) and H@1/3/10. MRR is the mean of the inverse of ranks of test cases and H@1/3/10 are the proportions of test cases that are ranked in the top 1/3/10, respectively. In the computation of ranks, we use the \textit{filtered} setting, i.e. we filter out the valid triplets that appeared in train, validation, and test sets among candidates.

\begin{wraptable}{r}{0.55\textwidth}
    \scriptsize
    \centering
    \begin{tabular}{c||cccc}
    \hline
    Dataset & YAGO & WIKI & ICEWS18 & GDELT \\ 
    \hline
    Entities & 10,623 & 12,544 & 23,033 & 7,691 \\
    Relations & 10 & 24 & 256 & 240 \\ \hline
    Training & 161,540 & 539,286 & 373,018 & 1,734,399 \\
    Validation & 19,523 & 67,538 & 45,995 & 238,765 \\
    Test & 20,026 & 63,110 & 49,545 & 305,241 \\ \hline
    Time gap & 1 year & 1 year & 1 day & 15 mins \\
    Timestamps & 189 & 232 & 304 & 2,751 \\ \hline
    Hier. score & 0.898 & 0.976 & 0.842 & 0.780
    \end{tabular}
    \caption{Statistics of four datasets. The last row gives the average of Krackhardt hierarchical scores for each dataset.}
    \label{table:datastat}
    \vspace{-0.5cm}
\end{wraptable}

We compare our \model to a diverse set of recent methods for reasoning on static KGs and temporal KGs. Static KG embedding models include TransE \citep{bordes2013translating}, DistMult \citep{yang2014embedding}, ComplEx \citep{trouillon2016complex}, ConvE \citep{dettmers2018convolutional}, RotatE \citep{sun2018rotate}, RGCN-DistMult \citep{schlichtkrull2018modeling}, and CompGCN-DistMult \citep{Vashishth2020Composition-based}. For the static KG methods, we simply remove all the timestamps in datasets and compare the results. On the other hand, temporal KG embedding models include TTransE \citep{jiang2016towards}, HyTE \citep{dasgupta2018hyte}, TA-DistMult \citep{garcia-duran-etal-2018-learning}, RE-Net \citep{jin-etal-2020-recurrent}, CyGNet \citep{Zhu_Chen_Fan_Cheng_Zhang_2021}, SeDyT \citep{zhou2021sedyt} and HIP Network \citep{he2021hipn}. For the other temporal KG embedding models such as CluSTeR \citep{li-etal-2021-search}, RE-GCN \citep{li2021temporal}, TimeTraveler \citep{sun-etal-2021-timetraveler}, and Tango \citep{han-etal-2021-learning-neural}, they used time-aware evaluation metrics so direct comparison was unavailable. 

\begin{wrapfigure}{r}{0.5\textwidth}
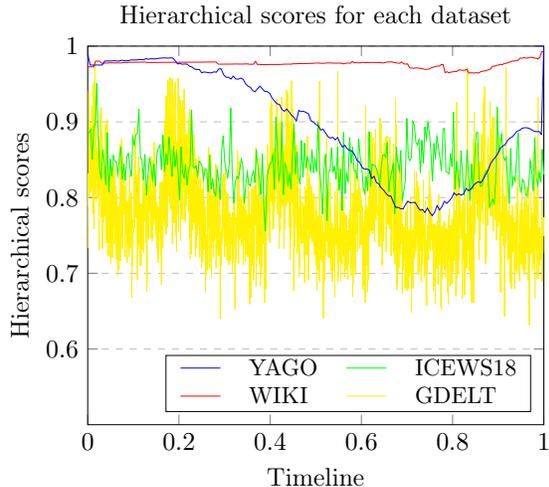

    \centering
    \vspace{-20pt}
    \resizebox{\linewidth}{!}{

    }
    \vspace{-1cm}
    \caption{Krackhardt hierarchical stores at each timestamp. The x-axis is the proportioned timestamps compared to the entire dataset.}
    \label{fig:hs}
    \vspace{0.2cm}
\end{wrapfigure}

\stitle{Results}
The hyperparameters were searched based on the MRR performance of validation sets. We used the Adam optimizer with a learning rate of $0.001$. The batch size was $1024$, and the training epoch was (maximum) $100$. The loss-controlling hyperparameter $\lambda$ was set to $0.01$. The embedding dimension was tuned among $100$, $150$, $200$, and $300$, and the model with dimension $200$ mostly outperformed.

\begin{table}[t!] \scriptsize
    \centering
    \setlength{\tabcolsep}{1pt}
    \resizebox{\textwidth}{!}{
    \begin{tabular}{l|cccc|cccc|cccc|cccc}
    \hline
    & \multicolumn{4}{c|}{YAGO}  & \multicolumn{4}{c|}{WIKI} & \multicolumn{4}{c|}{ICEWS18} & \multicolumn{4}{c}{GDELT} \\
Model      & MRR     & H@1   & H@3     & H@10   & MRR     & H@1   & H@3     & H@10    & MRR     & H@1   & H@3     & H@10   & MRR     & H@1  & H@3     & H@10       \\ \hline \hline
TransE      & 48.97 & 46.23 & 62.45 & 66.05 & 46.68 & 36.19 & 49.71 & 51.71 & 17.56 & 2.48 & 26.95 & 43.87 & 16.05 & 0.00 & 26.10 & 42.29 \\
DistMult    & 59.47 & 52.97 & 60.91 & 65.26 & 46.12 & 37.24 & 49.81 & 51.38 & 22.16 & 12.13 & 26.00 & 42.18 & 18.71 & 11.59 & 20.05 & 32.55 \\
ComplEx     & 61.29 & 54.88 & 62.28 & 66.82 & 47.84 & 38.15 & 50.08 & 51.39 & 30.09 & 21.88 & 34.15 & 45.96 & 22.77 & 15.77 & 24.05 & 36.33 \\
ConvE       & 62.32 & 56.19 & 63.97 & 65.60 & 47.57 & 38.76 & 50.10 & 50.53 & 36.67 & 28.51 & 39.80 & 50.69 & 35.99 & 27.05 & 39.32 & 49.44 \\
RotatE      & 65.09 & 57.13 & 65.67 & 66.16 & 50.67 & 40.88 & 50.71 & 50.88 & 23.10 & 14.33 & 27.61 & 38.72 & 22.33 & 16.68 & 23.89 & 32.29 \\
RGCN        & 41.30 & 32.56 & 44.44 & 52.68 & 37.57 & 28.15 & 39.66 & 41.90 & 23.19 & 16.36 & 25.34 & 36.48 & 23.31 & 17.24 & 24.96 & 34.36 \\
CompGCN     & 41.42 & 32.63 & 44.59 & 52.81 & 37.64 & 28.33 & 39.87 & 42.03 & 23.31 & 16.52 & 25.37 & 36.61 & 23.46 & 16.65 & 25.54 & 34.58 \\
\hline
TTransE     & 32.57 & 27.94 & 43.39 & 53.37 & 31.74 & 22.57 & 36.25 & 43.45 & 8.36  & 1.94  & 8.71  & 21.93 & 5.52  & 0.47  & 5.01  & 15.27 \\
HyTE        & 23.16 & 12.85 & 45.74 & 51.94 & 43.02 & 34.29 & 45.12 & 49.49 & 7.31  & 3.10  & 7.50  & 14.95 & 6.37  & 0.00  & 6.72  & 18.63 \\
TA-DistMult & 61.72 & 52.98 & 63.32 & 65.19 & 48.09 & 38.71 & 49.51 & 51.70 & 28.53 & 20.30 & 31.57 & 44.96 & 29.35 & 22.11 & 31.56 & 41.39 \\
\hline
RE-Net      & 65.16 & 63.29 & 65.63 & 68.08 & 51.97 & 48.01 & 52.07 & 53.91 & 42.93 & 36.19 & 45.47 & 55.80 & 40.42 & 32.43 & 43.30 & 53.70 \\
CyGNet      & 63.47 & 64.26 & 65.71 & 68.95 & 45.50 & 50.48 & 50.79 & 52.80 & \underline{46.69} & \underline{40.58} & \underline{49.82} & \underline{57.14} & 50.92 & \underline{44.53} & \underline{54.69} & \underline{60.99} \\
SeDyT-CONV  & 66.88 & $--$     & 67.05 & 68.73 & 52.90 & $--$     &52.96  & 54.00 & 45.91 & $--$     & 45.86 & 49.54 & \textbf{54.86}   & $--$     & 54.68 & 58.14 \\
HIP Network & \textbf{67.55} & \underline{66.32} & \textbf{68.49} & \textbf{70.37} & \textbf{54.71} & \textbf{53.82} & \textbf{54.73} & \textbf{56.46} & \textbf{48.37} & \textbf{43.51} & \textbf{51.32} & \textbf{58.49} & \underline{52.76} & \textbf{46.35} & \textbf{55.31} & \textbf{61.87} \\
\hline
\model      & \underline{67.52} & \textbf{66.46} & \underline{67.52} & \underline{69.28} & \underline{53.02} & \underline{51.98} & \underline{53.36} & \underline{54.55} & 41.38 & 34.21 & 44.25 & 55.17 & 40.08 & 32.98 & 42.84 & 53.26 \\ 
\hline
    \end{tabular}
    }
    \caption{Performances (in percentage) at temporal link prediction task on four datasets. The best results are boldfaced and the second best ones are underlined.}
    \label{table:results}
    \vspace{-1cm}
\end{table}

Table \ref{table:results} reports the link prediction result by \model and other methods on four datasets. The baseline results are adopted from \citet{Zhu_Chen_Fan_Cheng_Zhang_2021}, \citet{zhou2021sedyt}, and \citet{he2021hipn}. On the first two datasets, \model outperformed all the previous static and temporal KGs methods except for HIP Network \citep{he2021hipn}. Especially, \model outperformed the state-of-the-arts at H@1 on YAGO. In the WIKI, the most hierarchical dataset among four TKG datasets, \model relatively improved the performance by (maximum) $8.27\%$ (H@1 on WIKI) when compared to RE-Net \citep{jin-etal-2020-recurrent}, which can be considered as a Euclidean version of \model. Our method showed great performances on these datasets because, as we can see in Figure \ref{fig:hs} and Table \ref{table:datastat}, YAGO and WIKI are highly hierarchical data in general. 

\begin{wraptable}{r}{0.45\textwidth}
    \scriptsize
    \centering
    \vspace{-10pt}
    \setlength{\tabcolsep}{2.5pt}
    \begin{tabular}{l|cccc}
    \hline
    & \multicolumn{4}{c}{WIKI}\\
Model      & MRR     & H@1      & H@3     & H@10 \\
\hline \hline
\model       & \textbf{53.02} & \textbf{51.98} & \textbf{53.36} & \textbf{54.55}\\
HIP w/o hist. module       & 48.25 & 39.17 & 50.36 & 52.11 \\ \hline
    \end{tabular}
    \vspace{-10pt}
    \caption{Comparison of \model and HIP Network (SoTA) without the historical vocabulary module.}
    \label{table:comparisonwithSoTA}
    \vspace{-21pt}
\end{wraptable}

Experiments further revealed that our approach of implementing hyperbolic spaces strengthened the performance on the datasets with high hierarchical scores (i.e., WIKI, YAGO) whereas, with the other two datasets that have low hierarchical scores (i.e., GDELT and ICEWS18), ours did not outperform the earlier models.
Notably, the biggest difference between SoTA (HIP Network \cite{he2021hipn}) and ours is that the former deploys an additional module to deal with historical vocabulary while ours does not.
\citet{he2021hipn} provided the performance scores without the historical vocabulary module on WIKI, and the comparison showed that our model \model outperformed SoTA without the module (See Table \ref{table:comparisonwithSoTA}).
Furthermore, another model CyGNet \citep{Zhu_Chen_Fan_Cheng_Zhang_2021} which had better performance than ours on the datasets with low hierarchical scores also included a similar module that deals with historical vocabulary.
Given this, we speculate that the gap in the performance between ours and the earlier ones may be attributed to the implementation of such historical vocabulary modules.
In other words, adding a module of historical vocabulary to our model may improve the performance on these datasets regardless of hierarchical scores.

\begin{wraptable}{r}{0.5\textwidth}
    \scriptsize
    \centering
    \vspace{-10pt}
    \setlength{\tabcolsep}{2.5pt}
    \begin{tabular}{l|rrr|rrr}
    \hline
    & \multicolumn{3}{c|}{YAGO} & \multicolumn{3}{c}{WIKI}\\
Model      & MRR     & H@3     & H@10 & MRR     & H@3     & H@10\\
\hline \hline
\model (P)      & \textbf{67.52} & \textbf{67.52} & \textbf{69.28} & \textbf{53.02} & \textbf{53.36} & \textbf{54.55}\\
\model (L)      & 66.91 & 67.15 & 68.62 & 52.10 & 52.45 & 53.90\\ \hline
    \end{tabular}
    \vspace{-10pt}
    \caption{Comparison of \model embedded in \poincare model and \model embedded in Lorentz model.}
    \label{table:hyperbolic}
    \vspace{-21pt}
\end{wraptable}

Table \ref{table:hyperbolic} reports the performances of methods that are embedded in two different hyperbolic spaces: \model (P) refers to the one embedded in the \poincare model and \model (L) is the one in the Lorentz model. Generally, the model embedded in the \poincare model shows better performance.

\stitle{Ablation study}
As we mentioned, we present an ablation study about the contribution of time-varying curvature of graphs at single timestamps at local representation. We compare four models with different functions that describe the curvature: a learnable constant function, an additive time series as in Equation~\ref{timeseries}, a function of Krackhardt hierarchical scores as in Equation~\ref{hierscore}, and a combination of two variables as in Equation~\ref{combi}.

\begin{wraptable}{r}{0.5\textwidth}
    \scriptsize
    \centering
    \setlength{\tabcolsep}{2pt}
    \vspace{-10pt}
    \begin{tabular}{l|cccc}
    \hline
    & \multicolumn{4}{c}{YAGO} \\
Model      & MRR     & H@1   & H@3     & H@10 \\
\hline
\model w/ learnable const.      & 67.49 & \textbf{66.46} & 67.49 & 69.12\\
\model w/ time series      & \textbf{67.52} & \textbf{66.46} & \textbf{67.52} & \textbf{69.28} \\
\model w/ hierarchical score    & 67.49 & 65.89 & 67.11 & 68.61 \\
\model w/ both     & 66.79 & 65.57 & 67.10 & 68.54 \\ 
\hline
\hline
    & \multicolumn{4}{c}{WIKI} \\
Model      & MRR     & H@1   & H@3     & H@10 \\
\hline
\model w/ learnable const.       & \textbf{53.02} & \textbf{51.98} & \textbf{53.36} & \textbf{54.55} \\
\model w/ time series      & 52.51 & 51.32 & 52.85 & 54.10 \\
\model w/ hierarchical score    & 51.97 & 50.83 & 52.30 & 53.83 \\
\model w/ both     & 52.16 & 51.27 & 52.23 & 53.68 \\ 
\hline
    \end{tabular}
    \vspace{-15pt}
    \caption{Results of models with different types of learnable curvatures: a constant, a time series, a function of Krackhardt hierarchical scores, and a combination of a time series and hierarchical scores.}
    \label{table:ablation}
    \vspace{-1cm}
\end{wraptable}

From the results in Table \ref{table:ablation}, we observe that the model with an additive time series outperformed the other three models on YAGO dataset. The optimized global curvature at YAGO is $-2.367$, which means the global representation at each timestamp fits better in the hyperbolic space than in Euclidean space. Therefore, as we desired, the hierarchies derived from chronological properties are well represented in the hyperbolic space. On the other hand, the best local curvature at YAGO follows the additive time series model 
$$c_t = -\sigma \; [ \alpha * t + \beta * \sin (\omega * t) ],$$
where $\sigma$ is the ``Softplus'' function, $\alpha = -2.532 * 10^{-2}$, $\beta = -2.846 * 10^{-2}$, $\omega = -6.796 * 10^{-2}$. As $\alpha$ is negative, the curvature gets close to zero as the time increases. Interestingly, the performance got low when Krackhardt hierarchical scores were involved as an independent variable of curvature. 

However, as we see in the lower part of Table \ref{table:ablation}, the model with constant learnable curvature performed better in WIKI. This is because WIKI has rather consistent hierarchical structures compared to YAGO data. (See Figure \ref{fig:hs}.) Since the hierarchical score of WIKI is steady, the implementing time series or function of the hierarchical score may result in overfitting followed by lower performance.

\section{Conclusion}
In this paper, we proposed \model to tackle the extrapolation task on TKG reasoning in an autoregressive way. To address hierarchical relations within TKG, we relied on hyperbolic space rather than Euclidean space in both global and local representations. The global representation elaborates the hierarchies between knowledge graphs at different timestamps while local representation captures diverse hierarchical levels of knowledge graphs through the learnable curvature. According to the experimental results, \model performs great in the link prediction task on more hierarchical datasets. 

\section*{Acknowledgements}
The authors appreciate the reviewers and editors for their insightful comments and suggestions. 
Muhao Chen is supported by the National Science Foundation of United States grant IIS 2105329 and a faculty research award from Cisco.

\bibliography{custom,ma}

\begin{thebibliography}{61}
\providecommand{\natexlab}[1]{#1}
\providecommand{\url}[1]{\texttt{#1}}
\expandafter\ifx\csname urlstyle\endcsname\relax
  \providecommand{\doi}[1]{doi: #1}\else
  \providecommand{\doi}{doi: \begingroup \urlstyle{rm}\Url}\fi

\bibitem[Balazevic et~al.(2019)Balazevic, Allen, and
  Hospedales]{balazevic2019multi}
Ivana Balazevic, Carl Allen, and Timothy Hospedales.
\newblock Multi-relational poincar{\'e} graph embeddings.
\newblock \emph{Advances in Neural Information Processing Systems},
  32:\penalty0 4463--4473, 2019.

\bibitem[Beltrami(1868)]{beltrami1868teoria}
Eugenio Beltrami.
\newblock Teoria fondamentale degli spazii di curvatura costante.
\newblock \emph{Annali di Matematica Pura ed Applicata (1867-1897)}, 2\penalty0
  (1):\penalty0 232--255, 1868.

\bibitem[Bordes et~al.(2013)Bordes, Usunier, Garcia-Duran, Weston, and
  Yakhnenko]{bordes2013translating}
Antoine Bordes, Nicolas Usunier, Alberto Garcia-Duran, Jason Weston, and Oksana
  Yakhnenko.
\newblock Translating embeddings for modeling multi-relational data.
\newblock \emph{Advances in neural information processing systems}, 26, 2013.

\bibitem[Boschee et~al.(2015)Boschee, Lautenschlager, O’Brien, Shellman,
  Starz, and Ward]{boschee2015icews}
Elizabeth Boschee, Jennifer Lautenschlager, Sean O’Brien, Steve Shellman,
  James Starz, and Michael Ward.
\newblock Icews coded event data.
\newblock \emph{Harvard Dataverse}, 12, 2015.

\bibitem[Cannon et~al.(1997)Cannon, Floyd, Kenyon, Parry,
  et~al.]{cannon1997hyperbolic}
James~W Cannon, William~J Floyd, Richard Kenyon, Walter~R Parry, et~al.
\newblock Hyperbolic geometry.
\newblock \emph{Flavors of geometry}, 31:\penalty0 59--115, 1997.

\bibitem[Chami et~al.(2019{\natexlab{a}})Chami, Ying, R{\'{e}}, and
  Leskovec]{Chami2019HyperbolicGC}
Ines Chami, Zhitao Ying, Christopher R{\'{e}}, and Jure Leskovec.
\newblock Hyperbolic graph convolutional neural networks.
\newblock In Hanna~M. Wallach, Hugo Larochelle, Alina Beygelzimer, Florence
  d'Alch{\'{e}}{-}Buc, Emily~B. Fox, and Roman Garnett, editors, \emph{Advances
  in Neural Information Processing Systems 32: Annual Conference on Neural
  Information Processing Systems 2019, NeurIPS 2019, December 8-14, 2019,
  Vancouver, BC, Canada}, pages 4869--4880, 2019{\natexlab{a}}.

\bibitem[Chami et~al.(2019{\natexlab{b}})Chami, Ying, R{\'e}, and
  Leskovec]{chami2019hyperbolic}
Ines Chami, Zhitao Ying, Christopher R{\'e}, and Jure Leskovec.
\newblock Hyperbolic graph convolutional neural networks.
\newblock \emph{Advances in neural information processing systems}, 32,
  2019{\natexlab{b}}.

\bibitem[Chami et~al.(2020)Chami, Wolf, Juan, Sala, Ravi, and
  R{\'e}]{chami-etal-2020-low}
Ines Chami, Adva Wolf, Da-Cheng Juan, Frederic Sala, Sujith Ravi, and
  Christopher R{\'e}.
\newblock Low-dimensional hyperbolic knowledge graph embeddings.
\newblock In \emph{Proceedings of the 58th Annual Meeting of the Association
  for Computational Linguistics}, 2020.

\bibitem[Chen et~al.(2020)Chen, Huang, Xiao, Cai, and Jing]{chen2020hyperbolic}
Boli Chen, Xin Huang, Lin Xiao, Zixin Cai, and Liping Jing.
\newblock Hyperbolic interaction model for hierarchical multi-label
  classification.
\newblock In \emph{The Thirty-Fourth {AAAI} Conference on Artificial
  Intelligence, {AAAI} 2020, The Thirty-Second Innovative Applications of
  Artificial Intelligence Conference, {IAAI} 2020, The Tenth {AAAI} Symposium
  on Educational Advances in Artificial Intelligence, {EAAI} 2020, New York,
  NY, USA, February 7-12, 2020}, pages 7496--7503, 2020.

\bibitem[Chen and Quirk(2019)]{chen2019embedding}
Muhao Chen and Chris Quirk.
\newblock Embedding edge-attributed relational hierarchies.
\newblock In Benjamin Piwowarski, Max Chevalier, {\'{E}}ric Gaussier, Yoelle
  Maarek, Jian{-}Yun Nie, and Falk Scholer, editors, \emph{Proceedings of the
  42nd International {ACM} {SIGIR} Conference on Research and Development in
  Information Retrieval, {SIGIR} 2019, Paris, France, July 21-25, 2019}, pages
  873--876, 2019.

\bibitem[Dai et~al.(2021)Dai, Wu, Gao, and Jia]{Dai2021AHG}
Jindou Dai, Yuwei Wu, Zhi Gao, and Yunde Jia.
\newblock A hyperbolic-to-hyperbolic graph convolutional network.
\newblock 2021.

\bibitem[Dasgupta et~al.(2018)Dasgupta, Ray, and Talukdar]{dasgupta2018hyte}
Shib~Sankar Dasgupta, Swayambhu~Nath Ray, and Partha Talukdar.
\newblock Hyte: Hyperplane-based temporally aware knowledge graph embedding.
\newblock In \emph{Proceedings of the 2018 conference on empirical methods in
  natural language processing}, pages 2001--2011, 2018.

\bibitem[Deng et~al.(2020)Deng, Rangwala, and Ning]{deng2020dynamic}
Songgaojun Deng, Huzefa Rangwala, and Yue Ning.
\newblock Dynamic knowledge graph based multi-event forecasting.
\newblock In \emph{Proceedings of the 26th ACM SIGKDD International Conference
  on Knowledge Discovery \& Data Mining}, pages 1585--1595, 2020.

\bibitem[Dettmers et~al.(2018)Dettmers, Minervini, Stenetorp, and
  Riedel]{dettmers2018convolutional}
Tim Dettmers, Pasquale Minervini, Pontus Stenetorp, and Sebastian Riedel.
\newblock Convolutional 2d knowledge graph embeddings.
\newblock In \emph{Proceedings of the AAAI conference on artificial
  intelligence}, volume~32, 2018.

\bibitem[Ganea et~al.(2018{\natexlab{a}})Ganea, B{\'e}cigneul, and
  Hofmann]{ganea2018hyperbolic}
Octavian Ganea, Gary B{\'e}cigneul, and Thomas Hofmann.
\newblock Hyperbolic neural networks.
\newblock \emph{Advances in neural information processing systems}, 31,
  2018{\natexlab{a}}.

\bibitem[Ganea et~al.(2018{\natexlab{b}})Ganea, B{\'{e}}cigneul, and
  Hofmann]{Ganea2018HyperbolicNN}
Octavian{-}Eugen Ganea, Gary B{\'{e}}cigneul, and Thomas Hofmann.
\newblock Hyperbolic neural networks.
\newblock In Samy Bengio, Hanna~M. Wallach, Hugo Larochelle, Kristen Grauman,
  Nicol{\`{o}} Cesa{-}Bianchi, and Roman Garnett, editors, \emph{Advances in
  Neural Information Processing Systems 31: Annual Conference on Neural
  Information Processing Systems 2018, NeurIPS 2018, December 3-8, 2018,
  Montr{\'{e}}al, Canada}, pages 5350--5360, 2018{\natexlab{b}}.

\bibitem[Garc{\'\i}a-Dur{\'a}n et~al.(2018)Garc{\'\i}a-Dur{\'a}n,
  Duman{\v{c}}i{\'c}, and Niepert]{garcia-duran-etal-2018-learning}
Alberto Garc{\'\i}a-Dur{\'a}n, Sebastijan Duman{\v{c}}i{\'c}, and Mathias
  Niepert.
\newblock Learning sequence encoders for temporal knowledge graph completion.
\newblock In \emph{Proceedings of the 2018 Conference on Empirical Methods in
  Natural Language Processing}, pages 4816--4821, 2018.

\bibitem[Goel et~al.(2020)Goel, Kazemi, Brubaker, and
  Poupart]{goel2020diachronic}
Rishab Goel, Seyed~Mehran Kazemi, Marcus Brubaker, and Pascal Poupart.
\newblock Diachronic embedding for temporal knowledge graph completion.
\newblock In \emph{Proceedings of the AAAI Conference on Artificial
  Intelligence}, volume~34, pages 3988--3995, 2020.

\bibitem[G{\"{u}}l{\c{c}}ehre et~al.(2019)G{\"{u}}l{\c{c}}ehre, Denil,
  Malinowski, Razavi, Pascanu, Hermann, Battaglia, Bapst, Raposo, Santoro, and
  de~Freitas]{gulcehre2018hyperbolic}
{\c{C}}aglar G{\"{u}}l{\c{c}}ehre, Misha Denil, Mateusz Malinowski, Ali Razavi,
  Razvan Pascanu, Karl~Moritz Hermann, Peter~W. Battaglia, Victor Bapst, David
  Raposo, Adam Santoro, and Nando de~Freitas.
\newblock Hyperbolic attention networks.
\newblock In \emph{7th International Conference on Learning Representations,
  {ICLR} 2019, New Orleans, LA, USA, May 6-9, 2019}, 2019.

\bibitem[Han et~al.(2020)Han, Chen, Ma, and Tresp]{han2020dyernie}
Zhen Han, Peng Chen, Yunpu Ma, and Volker Tresp.
\newblock Dyernie: Dynamic evolution of riemannian manifold embeddings for
  temporal knowledge graph completion.
\newblock In \emph{Proceedings of the 2020 Conference on Empirical Methods in
  Natural Language Processing (EMNLP)}, pages 7301--7316, 2020.

\bibitem[Han et~al.(2021)Han, Ding, Ma, Gu, and
  Tresp]{han-etal-2021-learning-neural}
Zhen Han, Zifeng Ding, Yunpu Ma, Yujia Gu, and Volker Tresp.
\newblock Learning neural ordinary equations for forecasting future links on
  temporal knowledge graphs.
\newblock In \emph{Proceedings of the 2021 Conference on Empirical Methods in
  Natural Language Processing}, pages 8352--8364, 2021.

\bibitem[He et~al.(2021)He, Zhang, Liu, Liang, Zhang, and Zhang]{he2021hipn}
Yongquan He, Peng Zhang, Luchen Liu, Qi~Liang, Wenyuan Zhang, and Chuang Zhang.
\newblock Hip network: Historical information passing network for extrapolation
  reasoning on temporal knowledge graph.
\newblock In \emph{Proceedings of the Thirtieth International Joint Conference
  on Artificial Intelligence, {IJCAI-21}}, pages 1915--1921, 2021.

\bibitem[Huang et~al.(2019)Huang, Zhang, Li, and Li]{huang2019knowledge}
Xiao Huang, Jingyuan Zhang, Dingcheng Li, and Ping Li.
\newblock Knowledge graph embedding based question answering.
\newblock In \emph{Proceedings of the twelfth ACM international conference on
  web search and data mining}, pages 105--113, 2019.

\bibitem[Iversen and Birger(1992)]{iversen1992hyperbolic}
Birger Iversen and Iversen Birger.
\newblock \emph{Hyperbolic geometry}, volume~25.
\newblock 1992.

\bibitem[Jiang et~al.(2016)Jiang, Liu, Ge, Sha, Chang, Li, and
  Sui]{jiang2016towards}
Tingsong Jiang, Tianyu Liu, Tao Ge, Lei Sha, Baobao Chang, Sujian Li, and
  Zhifang Sui.
\newblock Towards time-aware knowledge graph completion.
\newblock In \emph{Proceedings of COLING 2016, the 26th International
  Conference on Computational Linguistics: Technical Papers}, pages 1715--1724,
  2016.

\bibitem[Jin et~al.(2020)Jin, Qu, Jin, and Ren]{jin-etal-2020-recurrent}
Woojeong Jin, Meng Qu, Xisen Jin, and Xiang Ren.
\newblock Recurrent event network: Autoregressive structure inferenceover
  temporal knowledge graphs.
\newblock In \emph{Proceedings of the 2020 Conference on Empirical Methods in
  Natural Language Processing (EMNLP)}, pages 6669--6683. Association for
  Computational Linguistics, 2020.

\bibitem[Krackhardt(2014)]{krackhardt2014graph}
David Krackhardt.
\newblock Graph theoretical dimensions of informal organizations.
\newblock In \emph{Computational organization theory}, pages 107--130.
  Psychology Press, 2014.

\bibitem[Leblay and Chekol(2018)]{leblay2018deriving}
Julien Leblay and Melisachew~Wudage Chekol.
\newblock Deriving validity time in knowledge graph.
\newblock In \emph{Companion Proceedings of the The Web Conference 2018}, pages
  1771--1776, 2018.

\bibitem[Leetaru and Schrodt(2013)]{leetaru2013gdelt}
Kalev Leetaru and Philip~A Schrodt.
\newblock Gdelt: Global data on events, location, and tone, 1979--2012.
\newblock In \emph{ISA annual convention}, volume~2, pages 1--49. Citeseer,
  2013.

\bibitem[Li et~al.(2021{\natexlab{a}})Li, Jin, Guan, Li, Guo, Wang, and
  Cheng]{li-etal-2021-search}
Zixuan Li, Xiaolong Jin, Saiping Guan, Wei Li, Jiafeng Guo, Yuanzhuo Wang, and
  Xueqi Cheng.
\newblock Search from history and reason for future: Two-stage reasoning on
  temporal knowledge graphs.
\newblock In \emph{Proceedings of the 59th Annual Meeting of the Association
  for Computational Linguistics and the 11th International Joint Conference on
  Natural Language Processing (Volume 1: Long Papers)}, pages 4732--4743,
  2021{\natexlab{a}}.

\bibitem[Li et~al.(2021{\natexlab{b}})Li, Jin, Li, Guan, Guo, Shen, Wang, and
  Cheng]{li2021temporal}
Zixuan Li, Xiaolong Jin, Wei Li, Saiping Guan, Jiafeng Guo, Huawei Shen,
  Yuanzhuo Wang, and Xueqi Cheng.
\newblock Temporal knowledge graph reasoning based on evolutional
  representation learning.
\newblock In \emph{Proceedings of the 44th International ACM SIGIR Conference
  on Research and Development in Information Retrieval}, pages 408--417,
  2021{\natexlab{b}}.

\bibitem[Liu et~al.(2019)Liu, Nickel, and Kiela]{liu2019hyperbolic}
Qi~Liu, Maximilian Nickel, and Douwe Kiela.
\newblock Hyperbolic graph neural networks.
\newblock \emph{Advances in Neural Information Processing Systems}, 32, 2019.

\bibitem[Liu et~al.(2018)Liu, Xiong, Sun, and Liu]{liu-etal-2018-entity}
Zhenghao Liu, Chenyan Xiong, Maosong Sun, and Zhiyuan Liu.
\newblock Entity-duet neural ranking: Understanding the role of knowledge graph
  semantics in neural information retrieval.
\newblock In \emph{Proceedings of the 56th Annual Meeting of the Association
  for Computational Linguistics (Volume 1: Long Papers)}, pages 2395--2405,
  2018.

\bibitem[L{\'o}pez and Strube(2020)]{lopez-strube-2020-fully}
Federico L{\'o}pez and Michael Strube.
\newblock A fully hyperbolic neural model for hierarchical multi-class
  classification.
\newblock In \emph{Findings of the Association for Computational Linguistics:
  EMNLP 2020}, pages 460--475, 2020.

\bibitem[Luo et~al.(2020)Luo, Zhang, Yang, Bo, Yang, Li, Qie, and
  Ye]{luo2020dynamic}
Wenjuan Luo, Han Zhang, Xiaodi Yang, Lin Bo, Xiaoqing Yang, Zang Li, Xiaohu
  Qie, and Jieping Ye.
\newblock Dynamic heterogeneous graph neural network for real-time event
  prediction.
\newblock In \emph{Proceedings of the 26th ACM SIGKDD International Conference
  on Knowledge Discovery \& Data Mining}, pages 3213--3223, 2020.

\bibitem[Ma et~al.(2021)Ma, Chen, Wu, and
  Peng]{ma-etal-2021-hyperexpan-taxonomy}
Mingyu~Derek Ma, Muhao Chen, Te-Lin Wu, and Nanyun Peng.
\newblock {H}yper{E}xpan: Taxonomy expansion with hyperbolic representation
  learning.
\newblock In \emph{Findings of the Association for Computational Linguistics:
  EMNLP 2021}, pages 4182--4194, 2021.

\bibitem[Mahdisoltani et~al.(2014)Mahdisoltani, Biega, and
  Suchanek]{mahdisoltani2014yago3}
Farzaneh Mahdisoltani, Joanna Biega, and Fabian Suchanek.
\newblock Yago3: A knowledge base from multilingual wikipedias.
\newblock In \emph{7th biennial conference on innovative data systems
  research}. CIDR Conference, 2014.

\bibitem[Montella et~al.(2021)Montella, Rojas~Barahona, and
  Heinecke]{montella-etal-2021-hyperbolic}
Sebastien Montella, Lina~M. Rojas~Barahona, and Johannes Heinecke.
\newblock Hyperbolic temporal knowledge graph embeddings with relational and
  time curvatures.
\newblock In \emph{Findings of the Association for Computational Linguistics:
  ACL-IJCNLP 2021}, pages 3296--3308, 2021.

\bibitem[Nickel and Kiela(2017)]{nickel2017poincare}
Maximilian Nickel and Douwe Kiela.
\newblock Poincar{\'{e}} embeddings for learning hierarchical representations.
\newblock In Isabelle Guyon, Ulrike von Luxburg, Samy Bengio, Hanna~M. Wallach,
  Rob Fergus, S.~V.~N. Vishwanathan, and Roman Garnett, editors, \emph{Advances
  in Neural Information Processing Systems 30: Annual Conference on Neural
  Information Processing Systems 2017, December 4-9, 2017, Long Beach, CA,
  {USA}}, pages 6338--6347, 2017.

\bibitem[Ren et~al.(2019)Ren, Chen, Li, Ren, Ma, and
  De~Rijke]{ren2019repeatnet}
Pengjie Ren, Zhumin Chen, Jing Li, Zhaochun Ren, Jun Ma, and Maarten De~Rijke.
\newblock Repeatnet: A repeat aware neural recommendation machine for
  session-based recommendation.
\newblock In \emph{Proceedings of the AAAI Conference on Artificial
  Intelligence}, volume~33, pages 4806--4813, 2019.

\bibitem[Sadeghian et~al.(2016)Sadeghian, Rodriguez, Wang, and
  Colas]{sadeghian2016temporal}
Ali Sadeghian, Miguel Rodriguez, Daisy~Zhe Wang, and Anthony Colas.
\newblock Temporal reasoning over event knowledge graphs.
\newblock In \emph{Workshop on Knowledge Base Construction, Reasoning and
  Mining}, 2016.

\bibitem[Schlichtkrull et~al.(2018)Schlichtkrull, Kipf, Bloem, van den Berg,
  Titov, and Welling]{schlichtkrull2018modeling}
Michael Schlichtkrull, Thomas~N. Kipf, Peter Bloem, Rianne van den Berg, Ivan
  Titov, and Max Welling.
\newblock Modeling relational data with graph convolutional networks.
\newblock In \emph{The Semantic Web}, pages 593--607. Springer International
  Publishing, 2018.

\bibitem[Sun et~al.(2021)Sun, Zhong, Ma, Han, and
  He]{sun-etal-2021-timetraveler}
Haohai Sun, Jialun Zhong, Yunpu Ma, Zhen Han, and Kun He.
\newblock {T}ime{T}raveler: Reinforcement learning for temporal knowledge graph
  forecasting.
\newblock In \emph{Proceedings of the 2021 Conference on Empirical Methods in
  Natural Language Processing}, pages 8306--8319, 2021.

\bibitem[Sun et~al.(2020)Sun, Chen, Hu, Wang, Dai, and Zhang]{sun2020knowledge}
Zequn Sun, Muhao Chen, Wei Hu, Chengming Wang, Jian Dai, and Wei Zhang.
\newblock Knowledge association with hyperbolic knowledge graph embeddings.
\newblock In \emph{Proceedings of the 2020 Conference on Empirical Methods in
  Natural Language Processing (EMNLP)}, pages 5704--5716, 2020.

\bibitem[Sun et~al.(2019)Sun, Deng, Nie, and Tang]{sun2018rotate}
Zhiqing Sun, Zhi-Hong Deng, Jian-Yun Nie, and Jian Tang.
\newblock Rotate: Knowledge graph embedding by relational rotation in complex
  space.
\newblock In \emph{International Conference on Learning Representations}, 2019.

\bibitem[Sur{\'\i}s et~al.(2021)Sur{\'\i}s, Liu, and
  Vondrick]{suris2021hyperfuture}
D{\'\i}dac Sur{\'\i}s, Ruoshi Liu, and Carl Vondrick.
\newblock Learning the predictability of the future.
\newblock In \emph{Proceedings of the IEEE/CVF Conference on Computer Vision
  and Pattern Recognition}, pages 12607--12617, 2021.

\bibitem[Trivedi et~al.(2017)Trivedi, Dai, Wang, and Song]{trivedi2017know}
Rakshit Trivedi, Hanjun Dai, Yichen Wang, and Le~Song.
\newblock Know-evolve: Deep temporal reasoning for dynamic knowledge graphs.
\newblock In \emph{international conference on machine learning}, pages
  3462--3471. PMLR, 2017.

\bibitem[Trivedi et~al.(2019)Trivedi, Farajtabar, Biswal, and
  Zha]{trivedi2019dyrep}
Rakshit Trivedi, Mehrdad Farajtabar, Prasenjeet Biswal, and Hongyuan Zha.
\newblock Dyrep: Learning representations over dynamic graphs.
\newblock In \emph{International conference on learning representations}, 2019.

\bibitem[Trouillon et~al.(2016)Trouillon, Welbl, Riedel, Gaussier, and
  Bouchard]{trouillon2016complex}
Th{\'e}o Trouillon, Johannes Welbl, Sebastian Riedel, {\'E}ric Gaussier, and
  Guillaume Bouchard.
\newblock Complex embeddings for simple link prediction.
\newblock In \emph{International conference on machine learning}, pages
  2071--2080. PMLR, 2016.

\bibitem[Ungar(2001)]{ungar2001hyperbolic}
Abraham~A Ungar.
\newblock Hyperbolic trigonometry and its application in the poincar{\'e} ball
  model of hyperbolic geometry.
\newblock \emph{Computers \& Mathematics with Applications}, 41:\penalty0
  135--147, 2001.

\bibitem[Vashishth et~al.(2020)Vashishth, Sanyal, Nitin, and
  Talukdar]{Vashishth2020Composition-based}
Shikhar Vashishth, Soumya Sanyal, Vikram Nitin, and Partha Talukdar.
\newblock Composition-based multi-relational graph convolutional networks.
\newblock In \emph{International Conference on Learning Representations}, 2020.

\bibitem[Veličković et~al.(2018)Veličković, Cucurull, Casanova, Romero,
  Liò, and Bengio]{velickovic2018graph}
Petar Veličković, Guillem Cucurull, Arantxa Casanova, Adriana Romero, Pietro
  Liò, and Yoshua Bengio.
\newblock Graph attention networks.
\newblock In \emph{International Conference on Learning Representations}, 2018.

\bibitem[Wang et~al.(2018)Wang, Zhang, Xie, and Guo]{wang2018dkn}
Hongwei Wang, Fuzheng Zhang, Xing Xie, and Minyi Guo.
\newblock Dkn: Deep knowledge-aware network for news recommendation.
\newblock In \emph{Proceedings of the 2018 world wide web conference}, pages
  1835--1844, 2018.

\bibitem[Wang et~al.(2021)Wang, Wei, Nogueira~dos Santos, Wang, Nallapati,
  Arnold, Xiang, Yu, and Cruz]{wang2021mixed}
Shen Wang, Xiaokai Wei, Cicero~Nogueira Nogueira~dos Santos, Zhiguo Wang,
  Ramesh Nallapati, Andrew Arnold, Bing Xiang, Philip~S Yu, and Isabel~F Cruz.
\newblock Mixed-curvature multi-relational graph neural network for knowledge
  graph completion.
\newblock In \emph{Proceedings of the Web Conference 2021}, pages 1761--1771,
  2021.

\bibitem[Xu et~al.(2020)Xu, Nayyeri, Alkhoury, Yazdi, and
  Lehmann]{xu2020temporal}
Chenjin Xu, Mojtaba Nayyeri, Fouad Alkhoury, Hamed Yazdi, and Jens Lehmann.
\newblock Temporal knowledge graph completion based on time series gaussian
  embedding.
\newblock In \emph{The Semantic Web -- ISWC 2020}, pages 654--671, Cham, 2020.
  Springer International Publishing.
\newblock ISBN 978-3-030-62419-4.

\bibitem[Yang et~al.(2014)Yang, Yih, He, Gao, and Deng]{yang2014embedding}
Bishan Yang, Wen-tau Yih, Xiaodong He, Jianfeng Gao, and Li~Deng.
\newblock Embedding entities and relations for learning and inference in
  knowledge bases.
\newblock \emph{arXiv preprint arXiv:1412.6575}, 2014.

\bibitem[Zhang et~al.(2016)Zhang, Yuan, Lian, Xie, and
  Ma]{zhang2016collaborative}
Fuzheng Zhang, Nicholas~Jing Yuan, Defu Lian, Xing Xie, and Wei-Ying Ma.
\newblock Collaborative knowledge base embedding for recommender systems.
\newblock In \emph{Proceedings of the 22nd ACM SIGKDD international conference
  on knowledge discovery and data mining}, pages 353--362, 2016.

\bibitem[Zhang et~al.(2020)Zhang, Chen, Wang, Song, and
  Roth]{zhang2020analogous}
Hongming Zhang, Muhao Chen, Haoyu Wang, Yangqiu Song, and Dan Roth.
\newblock Analogous process structure induction for sub-event sequence
  prediction.
\newblock In \emph{Proceedings of the 2020 Conference on Empirical Methods in
  Natural Language Processing (EMNLP)}, pages 1541--1550, 2020.

\bibitem[Zhang et~al.(2021)Zhang, Wang, Shi, Liu, and
  Song]{zhang2021lorentzian}
Yiding Zhang, Xiao Wang, Chuan Shi, Nian Liu, and Guojie Song.
\newblock Lorentzian graph convolutional networks.
\newblock \emph{ArXiv preprint}, abs/2104.07477, 2021.

\bibitem[Zhou et~al.(2021)Zhou, Orme-Rogers, Kannan, and
  Prasanna]{zhou2021sedyt}
Hongkuan Zhou, James Orme-Rogers, Rajgopal Kannan, and Viktor Prasanna.
\newblock Sedyt: A general framework for multi-step event forecasting via
  sequence modeling on dynamic entity embeddings.
\newblock In \emph{Proceedings of the 30th ACM International Conference on
  Information \& Knowledge Management}, pages 3667--3671, 2021.

\bibitem[Zhu et~al.(2021)Zhu, Chen, Fan, Cheng, and
  Zhang]{Zhu_Chen_Fan_Cheng_Zhang_2021}
Cunchao Zhu, Muhao Chen, Changjun Fan, Guangquan Cheng, and Yan Zhang.
\newblock Learning from history: Modeling temporal knowledge graphs with
  sequential copy-generation networks.
\newblock \emph{Proceedings of the AAAI Conference on Artificial Intelligence},
  35\penalty0 (5):\penalty0 4732--4740, 2021.

\end{thebibliography}
\bibliographystyle{plainnat}

\clearpage
\appendix
\section{Appendix}
\subsection{Basic Operations: Addition and Multiplication}
\label{appen:hyperoperations}

We first introduce basic addition and multiplication operations commonly used in neural networks for both \poincare and Lorentz models. Unlike in a Euclidean space, the addition of two vertex vectors in hyperbolic space is different from axis-wise addition. In \poincare model, we use M\"obius addition $\oplus_c$ for $\mathbf{x}, \mathbf{y} \in \mathbb{B}$ follows
\begin{equation*}
\mathbf{x} \oplus_c \mathbf{y} := \frac{(1 - 2 c \mathbf{x} \cdot \mathbf{y} - c \| \mathbf{y} \|^2) \mathbf{x} + (1 + c \| \mathbf{x}\|^2) \mathbf{y}}{1 - 2 c \mathbf{x} \cdot \mathbf{y} + c^2 \| \mathbf{x} \|^2 \| \mathbf{y} \|^2}
\end{equation*}
for addition operation where $\cdot$ is the dot product of two vectors~\cite{ungar2001hyperbolic,Ganea2018HyperbolicNN,gulcehre2018hyperbolic}. Note that as $c$ goes to $0$, the M\"obius addition converges to normal addition in Euclidean space. For multiplication, M\"obius matrix-vector multiplication $\otimes_c$ is defined as 

\begin{equation*}
M \otimes_c \mathbf{x} := (1 / \sqrt{c})\tanh \left(\frac{\|M \mathbf{x}\|}{\|\mathbf{x}\|} \tanh ^{-1}(\sqrt{c}\|\mathbf{x}\|)\right) \frac{M \mathbf{x}}{\|M \mathbf{x}\|}.
\end{equation*}

In the Lorentz model, we perform addition and multiplication via the tangent space. The logarithmic operation transforms vectors from $\mathbb{L}_c^d$ to the Euclidean tangent space $\mathbf{T}_x\mathbb{L}_c^d$ associated with the point $x \in \mathbb{L}_c^d$ and exponential operation conducts the reversed transformation~\cite{ma-etal-2021-hyperexpan-taxonomy}. Given $c = -1 / K(K>0)$, $\left<., .\right>_\mathcal{L}$ as the Minkowski inner product and $d_{\mathcal{L}}^{K}(x, y)=\sqrt{K} \operatorname{arcosh}\left(-\langle x, y \rangle_{\mathcal{L}} / K\right)$, they are defined as:

\begin{equation*}
\exp _{x}^{K}(v)=\cosh \left(\frac{\|v\|_{\mathcal{L}}}{\sqrt{K}}\right) x+\sqrt{K} \sinh \left(\frac{\|v\|_{\mathcal{L}}}{\sqrt{K}}\right) \frac{v}{\|v\|_{\mathcal{L}}},
\log _{x}^{K}(y)=d_{\mathcal{L}}^{K}(x, y) \frac{y+\frac{1}{K}\langle x, y\rangle_{\mathcal{L}} x}{\left\|y+\frac{1}{K}\langle x, y\rangle_{\mathcal{L}} x\right\|_{\mathcal{L}}}.
\end{equation*}
Hence, we could define matrix addition and multiplication on the Lorentz model by setting $x$ to to the origin point $\mathbf{o}$ if $P_{\mathbf{o} \rightarrow \mathbf{x}}^{K}(\cdot)$ is the parallel transport from $\mathbf{T}_\mathbf{o}\mathbb{L}_c^d$ to $\mathbf{T}_x\mathbb{L}_c^d$~\cite{Chami2019HyperbolicGC}:
\begin{equation*}
M \otimes^{K} \mathbf{x}:=\exp _{\mathbf{o}}^{K}\left(M \log _{\mathbf{o}}^{K}\left(\mathbf{x}\right)\right),
\mathbf{x} \oplus^{K} \mathbf{y}:=\exp _{\mathbf{x}}^{K}\left(P_{\mathbf{o} \rightarrow \mathbf{x}}^{K}(\mathbf{y})\right).
\end{equation*}

\subsection{Krackhardt Hierarchical Scores}
\label{appen:hierscore}
The Krackhardt hierarchical score measures how hierarchical the graph is as follows:
$$ Khs_{G} = \frac{\sum^{n}_{i,j=1} R_{i,j}(1-R_{j,i}) }{\sum^{n}_{i,j=1} R_{i,j}},$$
where $R$ is the adjacency matrix, i.e. $R_{i,j} = 1$ if there is an edge from node $i$ to $j$ and $0$ otherwise. For example, a graph full of symmetric relations has $Khs_G = 0$ while a tree (a graph with no symmetric relation) has $Khs = 1$. See \citet{krackhardt2014graph} for more details. 

\subsection{TKGs Datasets}
\label{appen:datasets}
The first dataset is a part of the dataset Global Databases of Events, Language, and Tone, or GDELT, and collected from 1/1/2018 to 1/31/2018 with a time interval of 15 minutes. ICEWS18 is from daily-event-based TKG Integrated Crisis Early Warning System (ICEWS), extracted from 1/1/2018 to 10/31/2018. The last two datasets are subsets of Wikipedia history and YAGO3, respectively. They both contain temporal fact with time frames in the form of $(s, r, o, [t_s, t_e])$, where $t_s$ is the starting time and $t_e$ is the ending time. Following the prior works \citep{jin-etal-2020-recurrent, Zhu_Chen_Fan_Cheng_Zhang_2021}, we divide the temporal facts (WIKI and YAGO) into timestamps with a time interval of one year.

\end{document}